\documentclass[lettersize,journal]{IEEEtran}
\usepackage{graphicx}
\usepackage{textcomp}
\usepackage{xcolor}
\usepackage{colortbl}
\definecolor{mygray}{gray}{0.98}
\definecolor{mygray1}{gray}{0.93}
\usepackage{algorithm}  
\usepackage{algpseudocode}  
\usepackage{amsmath}  
\usepackage{amsfonts}
\usepackage{amssymb}
\usepackage{latexsym}
\usepackage{caption, subcaption}
\usepackage{enumitem}
\usepackage{booktabs}
\usepackage{listings}
\usepackage{chngpage}
\usepackage{flushend}
\usepackage{makecell}
\usepackage{adjustbox}
\usepackage{bbding}
\usepackage{utfsym}
\usepackage{autobreak}
\usepackage{multirow}
\usepackage{siunitx}
\usepackage{textcomp,booktabs}
\usepackage{wasysym}

\usepackage{ragged2e}
\definecolor{seagreen}{rgb}{0.18, 0.55, 0.34}
\definecolor{royalpurple}{rgb}{0.47,0.32,0.66}
\definecolor{brown(traditional)}{rgb}{0.59, 0.29, 0.0}
\definecolor{blue}{rgb}{0.3, 0.2, 0.9}
\usepackage[colorlinks,
            linkcolor=blue,
            anchorcolor=blue,
            citecolor=blue]{hyperref}

\begin{document}


\title{ICST-DNET: An Interpretable Causal Spatio-Temporal Diffusion Network for Traffic Speed Prediction}

\author{Yi Rong, Yingchi Mao, Yinqiu Liu,~\IEEEmembership{Student Member,~IEEE}, Ling Chen, Xiaoming He,~\IEEEmembership{Member,~IEEE}, \\ and Dusit Niyato,~\IEEEmembership{Fellow,~IEEE} 
\thanks{Y. Rong, Y. Mao, and L. Chen are with the College of Computer and Software, Hohai University, Nanjing 210098, China (e-mail: rongyi1220@163.com; yingchimao@hhu.edu.cn; 221307050018@hhu.edu.cn)}
\thanks{Y. Liu and D. Niyato are with the School of Computer Science and Engineering, Nanyang Technological University, Singapore 639798 (e-mail: yinqiu001@e.ntu.edu.sg; dniyato@ntu.edu.sg)}
\thanks{X. He is with the college of Internet of Things, Nanjing University of Posts and Telecommunications, Nanjing 210003, China (e-mail: hexiaoming@njupt.edu.cn)}
\thanks{Y. Rong, Y. Liu, and X. He contributed equally to the work.}
\thanks{Corresponding author: Yingchi Mao.}
}

\markboth{}%
{Shell \MakeLowercase{\textit{et al.}}: A Sample Article Using IEEEtran.cls for IEEE Journals}


\maketitle

\begin{abstract}
Traffic speed prediction is significant for intelligent navigation and congestion alleviation.
However, making accurate predictions is challenging due to three factors: 
1) traffic diffusion, i.e., the spatial and temporal causality existing between the traffic conditions of multiple neighboring roads, 2) the poor interpretability of traffic data with complicated spatio-temporal correlations, and 3) the latent pattern of traffic speed fluctuations over time, such as morning and evening rush. Jointly considering these factors, in this paper, we present a novel architecture for traffic speed prediction, called \textit{Interpretable Causal Spatio-Temporal Diffusion Network} (ICST-DNET). 
Specifically, ICST-DENT consists of three parts, namely the Spatio-Temporal Causality Learning (STCL), Causal Graph Generation (CGG), and Speed Fluctuation Pattern Recognition (SFPR) modules. 
First, to model the traffic diffusion within road networks, an STCL module is proposed to capture both the temporal causality on each individual road and the spatial causality in each road pair. 
The CGG module is then developed based on STCL to enhance the interpretability of the traffic diffusion procedure from the temporal and spatial perspectives. 
Specifically, a time causality matrix is generated to explain the temporal causality between each road's historical and future traffic conditions. 
For spatial causality, we utilize causal graphs to visualize the diffusion process in road pairs. 
Finally, to adapt to traffic speed fluctuations in different scenarios, we design a personalized SFPR module to select the historical timesteps with strong influences for learning the pattern of traffic speed fluctuations. 
Extensive experimental results on two real-world traffic datasets prove that ICST-DNET can outperform all existing baselines, as evidenced by the higher prediction accuracy, ability to explain causality, and adaptability to different scenarios.
\end{abstract}


\section{Introduction}\label{s1}
\IEEEPARstart{T}{he} growing traffic demand has posed challenges for traffic managers. Accordingly, the Intelligent Transportation System (ITS) has received much attention, and traffic speed prediction, i.e., predicting the traffic speed of a given road at a given timestep, is of significant value in ITS \cite{dimitrakopoulos2010intelligent}. Accurate predictions made in advance can provide valuable information to traffic management authorities \cite{cao2020spectral}. Early attempts 

In the early stages, researchers utilized statistical methods, such as History Average (HA) \cite{liu2004summary} and shallow machine learning methods like Support Vector Regression (SVR) \cite{vanajakshi2004comparison}, to predict traffic speed. However, the modern traffic environment is complicated. Given a complex road network with multiple interconnected roads, the traffic condition of each road will affect its neighbors, called spatial dependency. Meanwhile, temporal dependency also exists since the historical traffic of each road significantly affects its future. These two dependency types influence each other, which we coin as spatio-temporal correlations. Moreover, both the spatial and temporal dependency changes over time, exhibiting great dynamics. Traditional statistical methods fail to capture such dynamic spatio-temporal correlations from the traffic data, resulting in low prediction accuracy. 

\begin{figure}[tp]
\centering
\includegraphics[width=0.6\textwidth, trim=140 5 50 10]{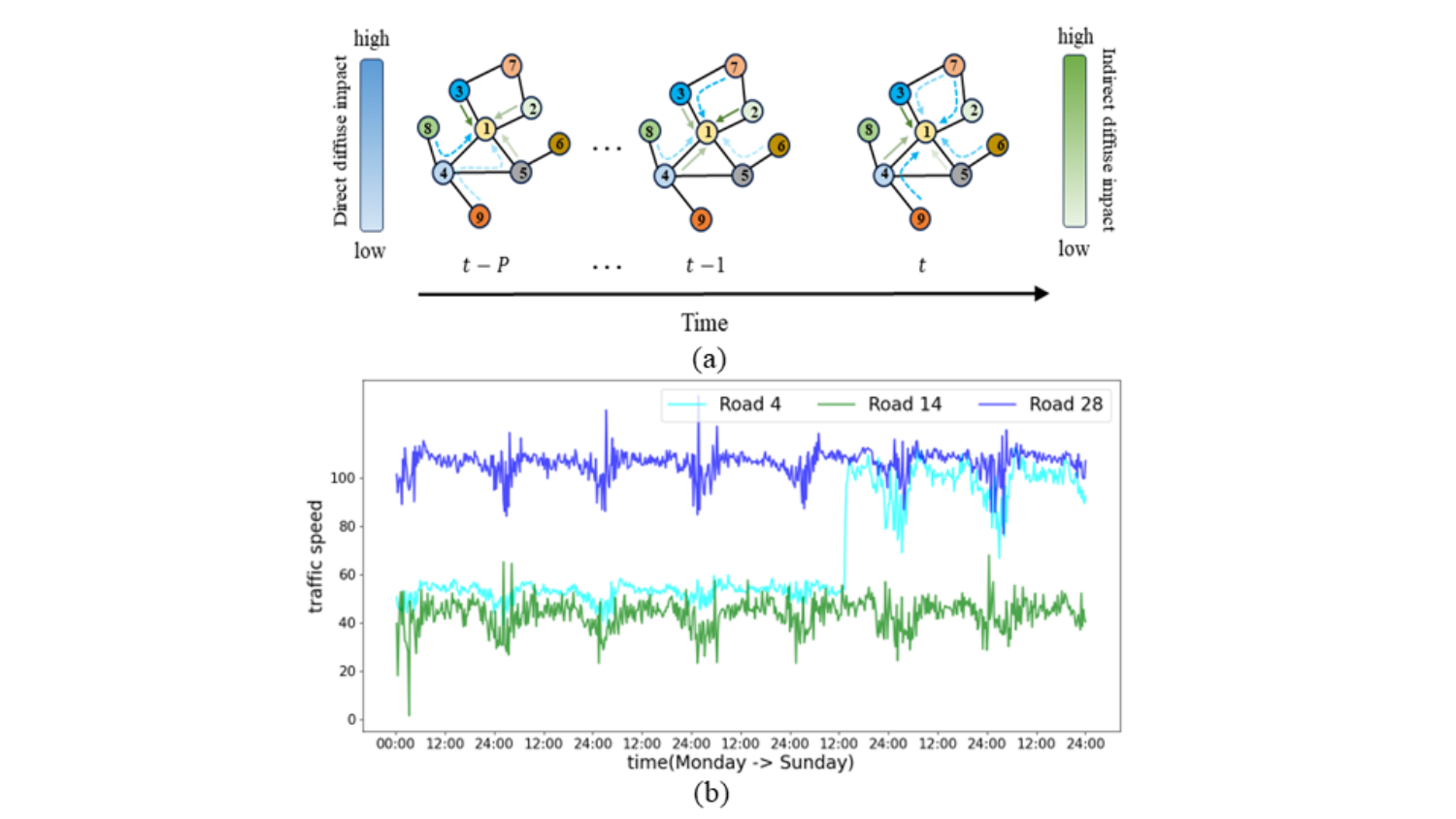}
\caption{\textbf{(a)} The traffic diffusion between Road 1 and its neighboring roads from timestep \textit{t-P} to \textit{t}. Solid black lines indicate the road network; solid green lines represent the direct diffusion between Road 1 and its first-order neighbors; blue dashed lines indicate the indirect diffusion between Road 1 and its second-order neighbors. Differences in the gradient color of the lines indicate different influence levels. \textbf{(b)} Examples of traffic speed fluctuation over time. Roads 14 and 28 are neighboring roads and exhibit similar patterns.}
\label{fig_1}
\end{figure}

Recently, deep learning methods with automatic feature extraction capabilities have gathered significant attention in traffic speed prediction. 
For instance, Diffusion Convolutional Recurrent Neural Network (DCRNN) \cite{li2017diffusion} and Spatio-Temporal Graph Convolutional Network (STGCN) \cite{yu2017spatio} adopted Graph Convolutional Network (GCN) to represent road networks and capture the spatial dependencies inside.
Meanwhile, the temporal dependencies are captured by time series networks, such as Recurrent Neural Network (RNN) or Temporal Convolutional Network (TCN). 
Nonetheless, these two proposals rely on local spatio-temporal representations and model road networks using fixed structures, limiting their capacity to explore dynamic spatio-temporal features. 
To this end, Graph WaveNet (GWN) \cite{wu2019graph} and Multivariate Time Graph Neural Network (MTGNN) \cite{wu2020connecting} employed adaptive graph modeling to represent the global spatial correlations within GCN and alleviated the inflexibility caused by the fixed structures. 
With the advancements of Transformer \cite{vaswani2017attention}, self-attention has been applied in extracting spatio-temporal correlations for accurate traffic speed prediction. 
For instance, Spatio-Temporal Graph Attention (ST-GRAT) \cite{park2020st} and Graph Multi-Attention Network (GMAN) \cite{zheng2020gman} enhanced the ability to represent the global and dynamic spatio-temporal correlations by substituting GCN and time series networks with spatial and temporal attention, respectively. 
Despite further improving the prediction accuracy, the existing methods ignore three important factors:

\textit{\textbf{Identification of Traffic Diffusion in Road Networks}}: For each road in a road network, its traffic condition is directly or indirectly affected by the neighboring roads over time, which we coin as the \textit{traffic diffusion} \cite{li2017diffusion, anwar2015roadrank, medina2022characterizing}. The essence of traffic diffusion is the reflection of spatio-temporal correlations from the causality perspective. As shown in Fig. \ref{fig_1}(a), the traffic condition of Road 1 is affected by the direct diffusion of its first-order neighbors (connected by green solid lines) and indirect diffusion of its higher-order neighbors (connected by blue dotted lines) through the network-wide propagation. Such direct and indirect diffusion is collectively referred to as spatial causality. Meanwhile, the current traffic conditions of Road 1 have a direct causal relationship with its history, which is called temporal causality. Such diffusion effects play a vital role in traffic prediction. Unfortunately, most existing works fail to explore the causality behind traffic diffusion.

\textit{\textbf{Interpretability of Advanced Spatio-Temporal Networks}}: Precisely representing dynamic spatio-temporal correlations is a prerequisite for accurate traffic speed prediction, while the resulting networks are hard to interpret. In this case, it is difficult to explain the rationale behind accurate predictions for improving decision-making in ITS. Although some studies intend to explain complex spatio-temporal networks \cite{bang2021explaining}, the human-friendly visualization of the traffic diffusion process is still unsolved, hindering us from building trustworthy ITS. 

\textit{\textbf{Adaptability to Traffic Speed Fluctuations in Different Scenario}}: In practice, traffic speed exhibits regular changes over time. As shown in Fig. \ref{fig_1}(b), Roads 4, 14, and 28 have clear morning and evening peaks. Likewise, the traffic speed on weekdays and weekends also varies. Such changes are called \textit{traffic speed fluctuation}, which should be considered in traffic speed prediction. The traffic speed fluctuation exists in both temporal and spatial dimensions. First, traffic speed simultaneously exhibits immediate and periodic changes over time, called short-term and long-term temporal dependencies. Furthermore, the spatial dependencies are also obvious, as the neighboring roads show similar patterns. 

To fill the research gap, this paper presents a novel architecture for traffic speed prediction called Interpretable Causal Spatio-Temporal Diffusion Network (ICST-DNET). Specifically, a Spatio-Temporal Causality Learning (STCL) module is first developed to learn both the temporal and spatial causality behind the traffic diffusion process. A Causal Graph Generation (CGG) module containing a time causality matrix and a series of causal graphs is then designed to visualize the captured causality, thereby enhancing the interpretability of traffic diffusion. Finally, a Speed Fluctuations Pattern Recognition (SFPR) module is developed to extract the spatio-temporal correlations related to traffic speed fluctuations. Particularly, we utilize a time filtering array to detect important historical timesteps that reveal the short-term and long-term temporal dependencies. Irrelevant historical timesteps are filtered out to reduce the computation costs. The main contributions of this paper are summarized as follows.

\begin{itemize}
\item We present a novel traffic prediction architecture with high accuracy and interpretability named ICST-DNET. To mine in-depth spatio-temporal correlations, we first design an STCL module, which explores temporal and spatial causality behind traffic diffusion using causal discovery approaches. By taking traffic diffusion into consideration, more spatial-temporal correlations can be revealed, facilitating accurate traffic speed prediction.

\item To enhance the interpretability of spatio-temporal networks for traffic speed prediction, we propose the CGG module with a time causality matrix and a series of causal graphs. The former can model the temporal causality between the past and prediction timesteps of the target roads. The latter aims to visualize the spatial causality of traffic conditions in road pairs.

\item To adapt to traffic speed fluctuations in different scenarios, we design the SFPR module to capture the short- and long-term temporal dependencies, as well as the spatial dependency. Particularly, a time filtering array is introduced, which aims to minimize the prediction noise by removing irrelevant historical timesteps according to empirical assumptions \cite{zheng2020gman, zou2023will}. In addition, we develop a Spatio-Temporal Fusion (ST-Fusion) method to adaptively incorporate temporal and spatial features through employing a gating mechanism to self-learn their respective impact weights.

\item Experimental results on two large-scale real-world traffic speed datasets, i.e., Ningxia-YC and METR-LA, prove that ICST-DNET is significantly superior to representative baselines. Specifically, ICST-DNET can reduce the errors in predicting traffic speeds for the next 12 horizons with 5-minute and 15-minute time slots among horizons. Meanwhile, the temporal causality in individual roads and spatial causality in road pairs can be detected. The traffic speed fluctuations in different scenarios are also accurately recognized.
\end{itemize}

The remainder of this paper is organized as follows. The related works regarding traffic speed prediction and causal discovery on time-series data are reviewed in Section \ref{s2}. Section \ref{s3} covers two main parts. The first part presents the preliminaries. The detailed design of ICST-DNET is introduced in the second part, including STCL, CGG, and SFPR modules. Experiments on two real-world traffic speed prediction datasets are shown in Section \ref{s4}. Finally, Section \ref{s5} concludes this paper.

\section{RELATED WORK}\label{s2}
In this section, we review the related works, including previous methods for traffic speed prediction, as well as the time series analysis based on causal discovery.

\subsection{Traffic Speed Prediction}
Traffic speed prediction has been explored for decades. Initially, researchers employed classical statistical methods like HA \cite{liu2004summary}. Unfortunately, these methods are unsuitable for traffic speed prediction due to weak nonlinear feature extraction capability. Shallow machine learning methods like SVR \cite{vanajakshi2004comparison} were then presented, with the goal of learning nonlinear correlations. However, these methods still fail to represent the in-depth spatial-temporal correlations. The primary reason lies in the manual feature selection and disjointed learning modules of these models, which are not suited for the context of massive traffic data.

\textit{\textbf{Road Network Modeling}}: With the flourishing development of deep learning in various fields \cite{forsyth2002computer, liddy2001natural, miglietta2023smart}, several attempts have been made to utilize this technique in traffic speed prediction. 
For instance, Qu et al. \cite{qu2021features} applied RNN and its variants, such as Long Short Term Memory (LSTM) \cite{hochreiter1997long} and Gated Recurrent Unit (GRU) \cite{cho2014properties}, to learn traffic speed pattern and perform short-term forecasting. 
However, they ignore the spatial dependency between neighboring roads. 
Some studies \cite{jo2018image} then modeled the road network by a standard form, e.g., a 2D matrix, and adopted Convolutional Neural Networks (CNNs) to extract the spatial dependency. 
Despite the advantages of modeling spatial relationships between neighboring roads through convolution operation, several key spatial information, such as dependencies in distant road pairs, cannot be effectively represented by standard forms, limiting the prediction accuracy.
In light of this, GCN \cite{zhou2020graph} achieves great attention since it employs graph theory to represent complex road networks as graphs, which can fully maintain the original spatial relationships of the road network.

\textit{\textbf{Representation of Spatio-Temporal Correlations}}: Due to these benefits, various proposals adopted GCN-based networks to capture spatio-temporal correlations in traffic speed data. 
For instance, DCRNN \cite{li2017diffusion} replaced linear operations in GRU by diffusion convolution \cite{atwood2016diffusion} and recurrently utilizes GCN to extract spatio-temporal correlations within each timestep. 
Likewise, STGCN \cite{yu2017spatio} combined GCN and TCN to extract spatial and temporal dependencies, respectively. 
We refer to such architectures as \textit{GCN + Time} (using TCN or RNN).
However, \textit{GCN + Time} methods are constrained by the adoption of local spatio-temporal features and predefined graph structures. 
To this end, models like GWN \cite{wu2019graph} and MTGNN \cite{wu2020connecting} extracted global spatio-temporal correlations using adaptive graphs and alleviated errors from predefined structures. 
With the rise of Transformers, self-attention mechanisms exhibit great potential in modeling spatio-temporal correlations. 
For instance, ST-GRAT \cite{park2020st} incorporated a stack of spatial and temporal attention into the vanilla Transformer architecture to predict the road speed of dynamically changing.
GMAN \cite{zheng2020gman} employed parallel spatial and temporal attention to predict the long-term traffic speed. 
Compared with \textit{GCN + Time}, the Transformer-based network further improves the capacity to extract global and dynamic spatio-temporal features. 
Nevertheless, these methods still suffer from three flaws.
First, as shown in Fig. 1, they ignore the effect caused by traffic diffusion in the complex road networks. 
Second, the models of the existing methods are less interpretable since complicated spatio-temporal correlations are accommodated. 
Third, the unnecessary historical timesteps fail to be filtered in existing methods, reducing the adaptivity of the traffic speed prediction model to fit the traffic speed fluctuations in different scenarios.

\subsection{Causal Discovery on Time Series}
\textit{\textbf{Granger Causality}}: Granger causality aims to detect the causal relationship between two time series \cite{granger1969investigating}. 
For example, Bressler et al. constructed a Vector Autoregressive (VAR) model using two time series and then performed an F-test on one time series while considering the presence or absence of the other one \cite{bressler2011wiener}. 
Since this method only considers two time series and ignores the influence of other series, it suffers from a high false alarm rate. 
Subsequent research extended Granger causality to pairwise conditions, accommodating all $m$ time series into one VAR model \cite{siggiridou2015granger}. 
However, such approaches suffer from two problems. 
First, detecting the causal relationship between any two series is by F-test, which may not align with the actual distribution of the traffic data.
Second, VAR models are linear and are only capable of learning simple causality. 
In contrast, our STCL and CGG modules do not rely on any distribution assumptions and have the capability to detect nonlinear causality.

\textit{\textbf{Graph Learning}}: Some researchers mine causality by embedding ridge regularization or lasso into VAR \cite{bahadori2013examination, cheng2014fblg}. Then, the strength of the causality between two time series can be deduced by evaluating the cumulative weights assigned to these series over all timesteps. If the sum of the weights tends to zero, the weaker the causal relationship between these two series, and vice versa. These models can create causal graphs in the form of adjacency matrices, which provide a more intuitive representation of causal relationships. However, they were still constrained by the linearity of the model.

\textit{\textbf{Supervised Learning-Based Causal Inference}}: Some approaches consider causal inference as a causal classification problem \cite{chikahara2018causal}, intending to assign a relationship label to two time series $X$ and $Y$, including 0 (no causal relationship), 1 ($X$ causes $Y$), and -1 ($Y$ causes $X$). This process can be accomplished in two steps: initially, using kernel mean embedding \cite{chikahara2018causal} maps the conditional distribution of $X$ to a point, considering both the previous states of $X$ when $Y$ was absent and present, respectively. Furthermore, it is necessary to compute the similarity as a metric for causality classification in the reproducing kernel Hilbert space. However, these methods require labeling relationships, which is a time-consuming and complex task. 

\textit{\textbf{LSTM}}: As a deep learning model for learning causal graphs, LSTM \cite{tank2021neural} evaluates the causality by examining whether the sum of weights in the initial layer of the network approaches zero. However, such a method exhibits poor performance due to the lack of a systematic learning process for different types of causality. Motivated by this, we present the STCL module, which takes a cascading approach to model temporal causality and generate causal graphs according to the captured causality.

\subsection{Motivation}
Overall, existing spatio-temporal networks are constrained by three factors in traffic speed prediction: a) They fail to take the effects of traffic diffusion into account. b) These methods have poor interpretability. c) The presence of unnecessary historical timesteps causes difficulties in fitting traffic speed fluctuations in multiple scenarios. Motivated by such fact, we present a novel traffic speed prediction model named ICST-DNET with STCL, CGG, and SFPR modules. Specifically, the STCL module leverages causal discovery to capture traffic diffusions from the temporal and spatial causality perspectives. Compared to previous causal discovery models, such as LSTM, the proposed method can extract temporal and spatial causality more systematically. CGG module is then proposed to visualize temporal and spatial causality, aiming to enhance the interpretability of ICST-DNET. Finally, the SFPR module is developed to improve the model's ability to fit traffic speed fluctuations in different scenarios, in which a time filtering array is introduced to remove irrelevant history timesteps.

\begin{figure*}[t]
    \centering
    \includegraphics[width=0.9\textwidth]{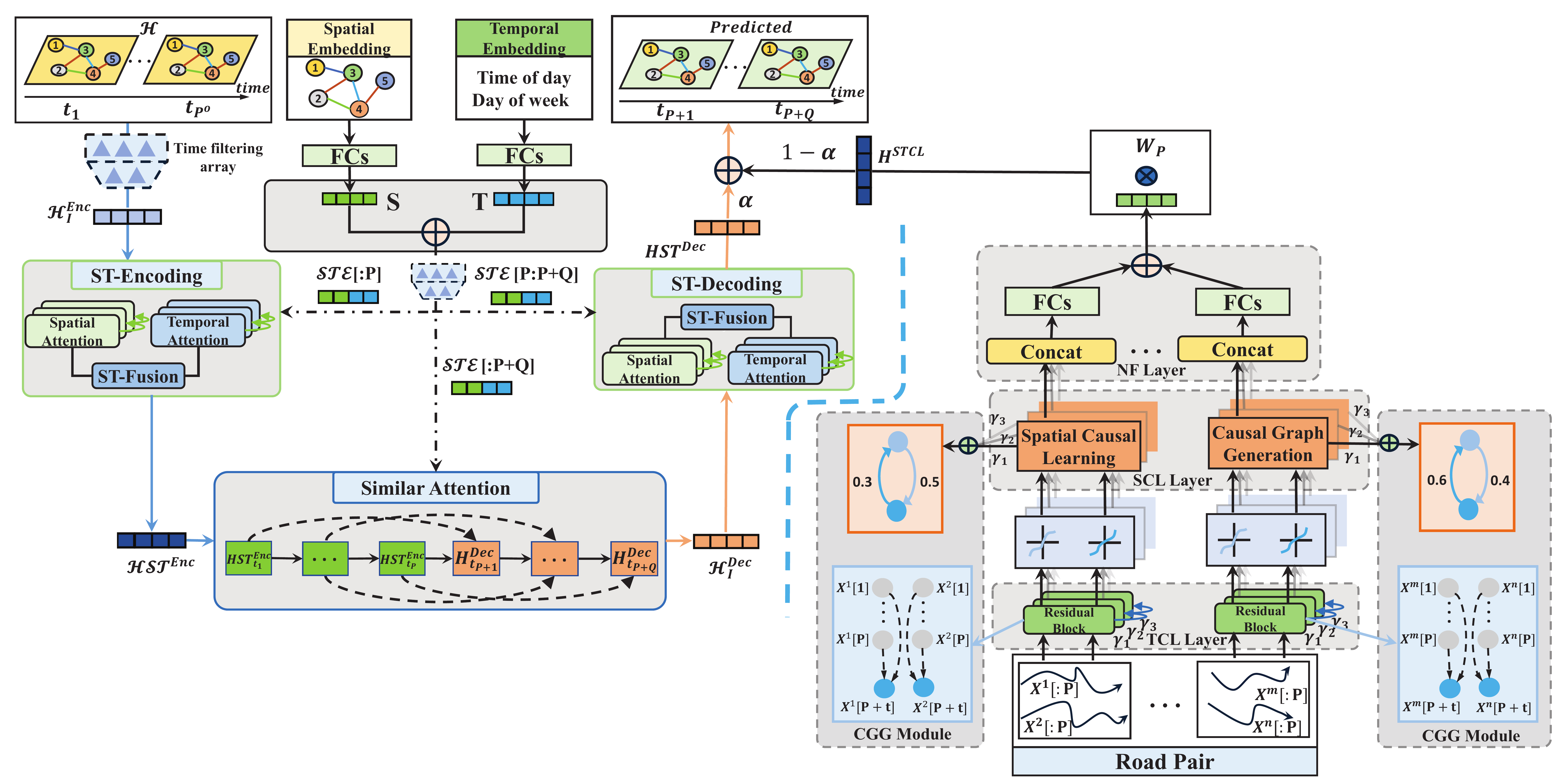}
    \caption{Overview of the proposed ICST-DNET. \textbf{Left:} SFPR module. $\mathcal{H}$ is the hidden state of the input. $\mathcal{H}_I^{E n c}$ is the hidden states selected by the time filtering array as the input of the ST-Encoding. $\mathcal{H S T}^{E n c}$ is the output of ST-Encoding. $\mathcal{H}_I^{D e c}$ is the input of the ST-Decoding, and $H S T^{D e c}$ denotes the production of the ST-Decoding. \textbf{Right:} STCL and CGG module. $X^i[: P]$ represents the time series of road $i$ in historical $P$ timesteps. $\gamma_j$ represents the learnable survival probability of the $j$th residual block.}
    \label{fig_2}
\end{figure*}

\begin{table}[]
\centering
\caption{Main notations.}
\begin{tabular}{ll}
\hline
Notations & Explanation \\ \hline
$A_{m n}$        &   {\makecell[{{p{6cm}}}]{\raggedright The global causal graph of the road pair $\left(x_m, x_n\right)$}}\\ 
$B_1$            &   {\makecell[{{p{6cm}}}]{\raggedright Time causality matrix}}  \\
$f(\cdot)$    & {\makecell[{{p{6cm}}}]{\raggedright The mapping function from historical traffic conditions to predictions}} \\
$G$       & The undirected graph generated by the entire road network         \\
$\mathcal{H}$    &   {\makecell[{{p{6cm}}}]{\raggedright Traffic speed embedding}}  \\
$\mathcal{H}_I^{\text {Dec }}$  &  {\makecell[{{p{6cm}}}]{\raggedright The output of similar attention}} \\
$\mathcal{H}_I^{E n c}$  & {\makecell[{{p{6cm}}}]{\raggedright The filtered traffic speed embedding}} \\
$H^{STCL}$       &   {\makecell[{{p{6cm}}}]{\raggedright The output of STCL module}}  \\
$H S T^{D e c}$  &  {\makecell[{{p{6cm}}}]{\raggedright The output of ST-Decoding}}  \\
$\mathcal{H S T}^{\text {Enc }}$   &  {\makecell[{{p{6cm}}}]{\raggedright The output of ST-Encoding}}  \\
$\mathcal{S T E}\left[: P^o\right]$  &    {\makecell[{{p{6cm}}}]{\raggedright Spatio-temporal embeddings of the entire road network for past $P^o$ timesteps}}  \\
$X$    & {\makecell[{{p{6cm}}}]{\raggedright The historical $P$ timesteps traffic condition of the entire road network $G$}}        \\
$Y$      & {\makecell[{{p{6cm}}}]{\raggedright The prediction sequences of future $Q$ timesteps for all roads in the road network}}       \\
$\Phi$    & The set of road pairs  \\
\hline
\end{tabular}
\label{symbols}
\end{table}

\section{ICST-DENT Design}\label{s3}
In this section, we illustrate the design of ICST-DENT.
Specifically, some key definitions (e.g., road network, traffic condition, and road pairs) are first given. Based on these definitions, we present the problem formulation of traffic speed forecasting. Second, a system overview is connected. Finally, we demonstrate the four modules of ICST-DENT in detail. For easy reference, the main notations are listed in TABLE \ref{symbols}.

\subsection{Preliminaries}
\textit{1) Definitions}:
In this part, we show several key definitions used in ICST-DENT.

\textit{\textbf{Definition I. Road network}}: In this paper, the road network is characterized as an undirected graph $G=(V, E)$. In the graph, the vertex set, denoted by $V$, contains $N=|V|$ roads; $E$ represents all the edges in $G$.

\textit{\textbf{Definition II. Traffic condition}}: At timestep $t_j$, $X^{t_j}=\left(x_{1,t_j}, x_{2,t_j}, \ldots, x_{N,t_j}\right) \in \mathbb{R}^{N \times C}$ represents the traffic condition of the entire road network $G$, with $x_{i,t_j}$ indicates the traffic condition of Road $i$ ($i \in \{1, 2, \dots, N\}$). Note that $C$ indicates the dimension of traffic condition, such as traffic flow and traffic speed.

\textit{\textbf{Definition III. Road pair}}: The set of road pairs is defined as $\left|\left(x_m, x_n\right)\right| \in \mathbb{R}^{P \times 2 \times C}$, where $m$ and $n$ $\in \{1, 2, \dots, N\}$. $P$ represents the length of the historical timesteps selected by the time filtering array, which is detailed below. Assuming that $x_m$ is the target road, any given road $x_n$ can establish a road pair $\left(x_m, x_n\right)$ with it only if they are first-order or second-order neighbors (see Fig. 1).

\textit{2) Problem Formulation}: Based on the above definition, the traffic speed prediction problem can be formally represented as follows. 

\textit{\textbf{Problem. Traffic speed prediction}}: Given the road network $G=(V, E)$, the observations of $P$ historical timesteps $X=\left(X^{t_1}, X^{t_2}, \ldots, X^{t_P}\right) \in \mathbb{R}^{P \times N \times C}$, and road pairs $\Phi=\left\{\left(x_1, x_2\right),\left(x_4, x_9\right), \ldots,\left(x_m, x_n\right)\right\},\left(x_m, x_n\right)\} \in \mathbb{R}^{P \times 2 \times C}$. The objective is to train a mapping function $f(\cdot)$ that forecasts the traffic speed of all roads in the road network with the next $Q$ timesteps:
\begin{equation}
(X, \Phi) \underset{f(\cdot)}{\rightarrow} Y,
\end{equation}
where $Y=\left(\hat{Y}^{t_{P+1}}, \hat{Y}^{t_{P+2}}, \ldots, \hat{Y}^{t_{P+Q}}\right) \in \mathbb{R}^{Q \times N \times C}$ is the predicted output. Note that since we focus on prediction traffic speed, $C$ is set to $1$ in this paper.

\subsection{System Overview}
As shown in Fig. \ref{fig_2}, to solve the above problem, we present the ICST-DNET system with three modules. 
First, the STCL is proposed, which mines the traffic condition of road pairs for capturing traffic diffusion in the entire road network from the temporal and spatial causality perspectives. 
Specifically, STCL follows the basic principles of Granger Causality \cite{granger1969investigating}, the most commonly employed causal discovery method. 
Moreover, we introduce deep neural networks to incorporate nonlinearity into Granger causality detection. 
Consequently, in contrast to traditional Granger Causality, STCL can not only capture the causality of historical timesteps to future timesteps in a road from the temporal dimension but also learn the spatial causality between two roads in a road pair.

We then propose a novel module based on a time-causality matrix and a causal graph named CGG to visualize the temporal and spatial causality, thereby explaining the traffic diffusion. 
Specifically, the temporal causality matrix is derived from the nonlinear function for learning temporal causality in STCL, in which the elements in matrices reflect useful causal information involving different historical timesteps to final regressions. The causal graph is constructed based on the spatial causality learning layer in STCL. Each node in the graph denotes a road. The edges are directed and treated intuitively as spatial causality between two connected roads. 

Last but not least, an intelligent module based on the time filtering array, spatial and temporal attention, ST-Fusion, and similar attention called SFPR module is present to adapt to traffic speed fluctuations in different scenarios. 
Specifically, considering that some irrelevant timesteps in the time window may bring noise, a time filtering array is applied to solve this issue by removing these timesteps based on empirical assumptions. Then, since spatio-temporal attention mechanisms are effective in dynamically adjusting the degree of model attention in both the temporal and spatial dimensions, enabling us to effectively handle the complex relationships of these two dimensions, we adopt it to extract spatial dependencies and short-term and long-term temporal dependencies. 
Afterward, since ST-Fusion facilitates adaptive fusion of spatio-temporal correlations, it is introduced to reasonably integrate temporal and spatial dependencies. 
Finally, similar attention is introduced to select the most relevant historical information for reducing error propagation in prediction.

\subsection{STCL Module Design}
As shown in the right part of Fig. \ref{fig_2}, the STCL module adopts a hierarchical architecture with three components, namely the Temporal Causality Learning (TCL) layer, the Spatial Causality Learning (SCL) Layer, and the Nonlinear Fusion (NF) layer. Below, we detail how each layer is applied and collaborates with each other to identify traffic diffusion from the causality perspective.

\begin{figure}[tpb]
\centering
\includegraphics[width=0.49\textwidth]{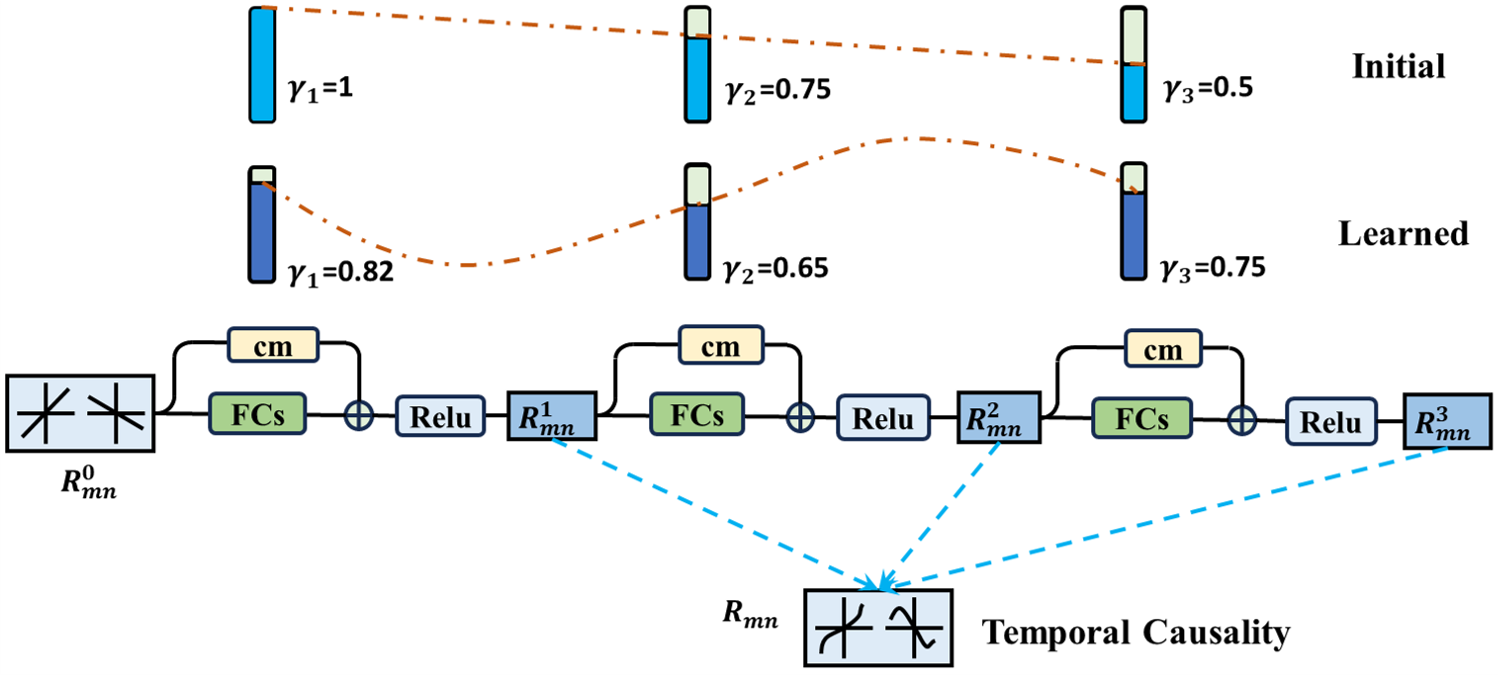}
\caption{The Temporal Causality Layer. We employ a ResNet model incorporating Learnable Layer Relevance $\gamma_q$ concerning the spatial causality. The outputs of all the residuals are forwarded to the SCL layer.}
\label{fig_3}
\end{figure}

\textit{1) Temporal Causality Learning (TCL)}: The TCL layer, implemented by Residual Neural Networks (ResNet) \cite{he2016deep}, is deployed to capture temporal causality within each road pair. The main reason for adopting the ResNet is that it can employ stacked residual blocks to represent different degrees of temporal causality, and the architecture is deep yet easily trainable \cite{xu2019scalable}. Specifically, the detailed architecture of TCL is displayed in Fig. \ref{fig_3}. By setting historical input sequences of road $m$ and $n$ $R_{m n}^0=\left(x_m, x_n\right)^T \in \mathbb{R}^{2 \times P}$, where $\left(x_m, x_n\right) \in \Phi$, the proposed ResNet is defined as: 
\begin{equation}
R_{m n}^q=\operatorname{ReLU}\left(\gamma_q\left(R_{m n}^{q-1} B_q\right)+\zeta\left(R_{m n}^q\right)\right),
\end{equation} where $R_{m n}^q \in \mathbb{R}^{2 \times d_{\text {model }}}$ ($q=\{1,2, \ldots, Q_R\}$) denotes the output of the $q$th residual block, which is the representation of a specific degree of temporal causality. The total number of residual blocks is denoted as $Q_R$, and $d_{\text {model }}$ is a hyperparameter of ICST-DNET. The weight matrix in the $q$th block is denoted as $B_q \in \mathbb{R}^{d_{\text {model }} \times d_{\text {model}}}$, $q=\{2,3, \ldots, Q_R\}$ and the first weight is $B_1 \in \mathbb{R}^{P \times d_{\text {model }}}$. $\zeta$ is a constant mapping \cite{he2016deep}. Generally, as ResNet delves deeper, there is more potential to capture in-depth temporal causality. Actually, it is hard to distinguish which block is more critical. Thus, $\gamma_q$ is the self-learning parameter to assign a weight to each residual block according to its importance and can be defined as follows:
\begin{equation}
\gamma_q=1-\frac{q}{Q_R}\left(1-\gamma_{Q_R}\right),
\end{equation} where $\gamma_{Q_R}$ is set as $0.5$ in our paper. This is because the highest prediction accuracy is found at $\gamma_{Q_R}=0.5$ after repeated attempts. The outputs from all $Q_R$ residuals $R_{m n}^{1: Q_R}$ approximate $R_{m n}$ and will be fed into the SCL layer.

\textit{2) Spatial Causality Learning (SCL)}: After receiving the temporal causality representations from $Q_R$ residual blocks, the SCL layer is applied to learn the spatial causality between two roads of each road pair. In detail, for each $R_{m n}^q \in R_{m n}$, the spatial causal matrix $A_{m n}^q \in \mathbb{R}^{2 \times 2}$ is assigned and the SCL layer can be expressed by
\begin{equation}
\begin{array}{c}
\left(\hat{R}_{m n}^q\right)^T=\left(R_{m n}^q\right)^T A_{m n}^q, \\
A_{m n}^q=\left[\begin{array}{ll}
a_{11} & a_{12} \\
a_{21} & a_{22}
\end{array}\right],
\end{array}
\end{equation} where $a_{12}$ and $a_{21}$ represent the degree of spatial causality of the road $m$ to $n$ and $n$ to $m$, respectively. While $a_{11}$ and $a_{22}$ stand for the degree of spatial causality of road $m$ and $n$ to themselves, respectively. $(\hat{R}_{m n}^q)^T$ represents the spatial causality representation of roads $m$ and $n$ under the output of the $q$th residual block. Both $(\hat{R}_{m n}^q)^T$ and $(R_{m n}^q)^T$ have dimensions of $\mathbb{R}^{d_{\text {model }} \times 2}$. Such an operation can be extended to the entire road network $G$.


\textit{3) Nonlinear Fusion (NF)}: To fuse the temporal and spatial causality of the entire road network and use it to denote traffic diffusion, we design the NF layer. Accordingly, the inputs to the NF layer are all $\hat{R}_{m n}^q \in \mathbb{R}^{2 \times d_{\text {model }},}, q=\{1,2, \ldots, Q\}$ that have undergone the operation of Eq. (4). By concatenating them column-wise, the entire temporal and spatial causality of a road pair $\left(x_m, x_n\right)$ is denoted as $\hat{R}_{m n}^{\gimel} \in \mathbb{R}^{2 \times \omega}$, where $\omega=Q_R \times d_{\text {model }}$ and means the dimensions obtained through concatenating. After receiving it, we leverage a multi-layer fully connected network (FC) to extract intervariable nonlinearity in the road pair. The output of the $s$th layer is defined as:
\begin{equation}
F_{m n}^s=\sigma\left(F_{m n}^{s-1} W_s+b_s\right), s=1,2, \ldots, D,
\end{equation} where $F_{m n}^{s-1} \in \mathbb{R}^{2 \times d_{\text {model }}}$ is the output of $(s-1)$th layer and set $F_{m n}^0=\hat{R}_{m n}^{\gimel}$. $W_s \in \mathbb{R}^{d_{\text {model }} \times d_{\text {model }}}$, $W_1 \in \mathbb{R}^{\omega \times d_{\text {model }}}$, and $b_s$ are the trainable weight matrices and bias vector, respectively. The activation function here is $tanh$. The intervariable nonlinearity representation is $F_{m n}^D \in \mathbb{R}^{2 \times d_{\text {model }}}$, where $D$ denotes the total number layers of the FC.

We sequentially remove each road pair from the $\Phi$ set and utilize Eqs. (2)-(5) to learn the spatial and temporal causality of each road pair. Afterward, to extend to the entire road network, additive operations are applied to aggregate the temporal and spatial causality of each road pair in the road network. The temporal and spatial causality across the entire road network is then represented as:
\begin{equation}
F^D=F_{12}^D + \cdots + F_{m n}^D,
\end{equation} where $F^D \in \mathbb{R}^{2 \times d_{\text {model }}}$ indicates the temporal and spatial causality of the road network, i.e., traffic diffusion.

Finally, as depicted in Fig. \ref{fig_2}, considering the feature fusion in the prediction layer of ICST-DNET, we transform $F^D$ into $\mathbb{R}^{1 \times N \times d_{\text {model }}}$ and then obtain the final traffic diffusion representation $H^{\text {STCL }} \in \mathbb{R}^{Q \times N \times d_{\text {model }}}$ by the following FC:
\begin{equation}
H^{S T C L}=F^D W_P,
\end{equation} where $W_P \in \mathbb{R}^{Q \times d_{\text {model }} \times d_{\text {model }}}$ denotes learnable parameters in the FC.

\begin{figure*}[t]
    \centering
    \includegraphics[width=0.7\textwidth]{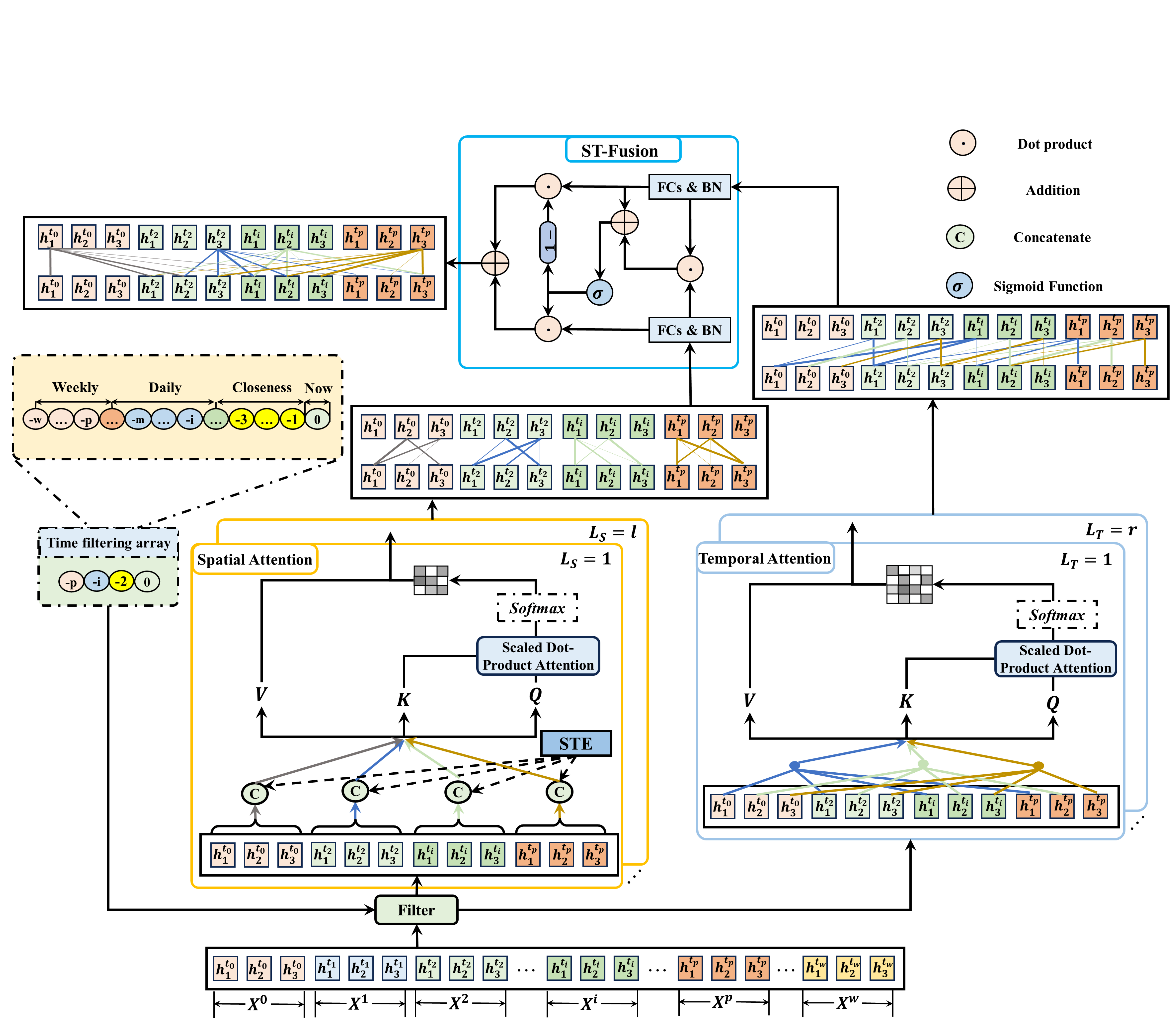}
    \caption{ST-Encoding. There are three steps: 1) select relevant spatio-temporal neighbors based on the time filtering array, 2) spatial and temporal attention is designed to extract dynamic spatio-temporal correlations, and 3) dynamic spatio-temporal properties are adaptively fused via ST-Fusion.}
    \label{fig_4}
\end{figure*}

\subsection{CGG Module Design} Apart from modeling traffic diffusion of the entire road network, a profound understanding of the specific diffusion processes is crucial to establishing a trustworthy ITS. In light of this, we propose the module named CGG to interpret traffic diffusion from the perspectives of temporal and spatial causality, assisted by time causality matrices and causal graphs. Next, we present the design details.

\textit{\textbf{Time Causality Matrix}}: First, we leverage a time causality matrix originating from TCL to represent the temporal causality existing in each road pair. Given a road pair $\left(x_m, x_n\right)$, the learnable weight matrix $B_1 \in \mathbb{R}^{P \times d_{\text {model }}}$ in the first residual block of TCL is retained as a time causality matrix. The main reason is that the matrix $B_1$ is composed of weights assigned to different historical timesteps, which can portray their causality on prediction. Each row represents the causality representations of a corresponding historical timestep for individual roads in $\left(x_m, x_n\right)$. If the sum of weight values in a row is larger, the corresponding historical timestep has a greater impact on the prediction, thus indicating a more significant causal relationship among them.

\textit{\textbf{Causal Graphs}}: Furthermore, to capture the spatial causality between two roads of each road pair, causal graphs are proposed. Assuming a road pair $\left(x_m, x_n\right)$, for $R_{m n}^q$ from the $q$th residual of TCL, we perform the preceding procedure (Eq.(4)) and acquire the spatial causal matrix $A_{m n}^q \in \mathbb{R}^{2 \times 2}$, which is regarded as a local causal graph. The elements in the matrix reflect the causality among roads $m$ and $n$. For instance, when the element value of road $m$ to $n$ is greater than the value of $n$ to $m$, we consider that road $m$ influences $n$, i.e., the direction of the connected edge is from $m$ to $n$. Since local causal graphs have different relevance to various degrees of temporal causality, it is necessary to develop a method to integrate different local causal graphs generated under different residuals. Therefore, we conduct a weighted sum of each local causal graph to produce a global causal graph. The global causal graph $A_{m n} \in \mathbb{R}^{2 \times 2}$ that reflects the spatial causality relationships between various roads in a road pair $\left(x_m, x_n\right)$, is defined as:
\begin{equation}
A_{m n}=\sum_{q=1}^{Q_R} \gamma_q A_{m n}^q,
\end{equation} where $\gamma_q$ is the learned relevance of $R_{m n}^q$.

After each road pair in the $\Phi$ is processed by CGG, the temporal and spatial causality of the entire road network is acquired as traffic diffusion. Section \ref{s4} shows the interpretation results by visualizing the time causality matrix and global causal graph to different road pairs.




\subsection{SFPR Module Design}
To overcome traffic speed fluctuations in multiple scenarios and thus improve traffic prediction accuracy, we develop the SFPR module.
Composed of five components, namely the feature embedding layer, time filtering layer, Spatio-Temporal Encoding (ST-Encoding), similar attention, and Spatio-Temporal Decoding (ST-Decoding), the SFPR module forms a pipeline architecture, as shown in the left part of Fig. \ref{fig_2}. 
Next, we discuss the details of each component and how they collaborate to adapt to traffic speed fluctuations.
Note that due to the frequent use of the nonlinear transformation function in SFPR, we first define it as follows:
\begin{equation}
f(x)=\operatorname{ReLU}(W x+b),
\end{equation} 
where $\operatorname{ReLU}$ refers to the activation function, $W$ and $b$ represent trainable weight and bias, respectively.

\textit{1) Feature Embedding Layer}: Since the neural networks cannot directly process the raw traffic data, it is necessary to transform different forms of input data, e.g., traffic speed and diffusion, into embeddings with uniform dimensions. The types of embeddings are described as follows.

\textit{\textbf{Traffic Speed Embedding}}: For the convenience of collaborative training with other data and obtaining expressive representations, it is indispensable to transform the traffic speed data of past $P^o$ timesteps $X$ into higher-dimensional features. Particularly, we apply an FC with two layers to convert dimensions from $C$ to $d_{\text {model}}$, which can be expressed as $\mathcal{H}=\left(H^{t_1}, H^{t_2}, \ldots, H^{t_{P^o}}\right) \in \mathbb{R}^{P^0 \times N \times d_{\text {model }}}$.

\textit{\textbf{Spatio-Temporal Embedding}}: Due to the influence of traffic diffusion on the evolution of traffic conditions, it is necessary to consider the road location information of roads when predicting traffic speed. Specifically, the \textit{node2vec} technique \cite{grover2016node2vec} is leveraged to learn road representations. The results are then fed into two-layer FCs to obtain spatial embeddings, denoted as $R^{N \times d_{\text {model}}}$.

The spatial embedding is insufficient because it only focuses on static representations and lacks dynamic correlations between roads in the road network. To circumvent this issue, a time embedding method is further designed to encode each timestep into a vector. To be specific, we adopt one-hot encoding to encode the time of one day and days of one week. The Time of one day is projected to $\mathbb{R}^{\mathrm{Z}}$, where $Z$ stands for the total number of timesteps in a day. The days of one week are projected to 7 slots (from Monday to Sunday). Then, the two projections are concatenated into a vector $\mathbb{R}^{\mathrm{7+Z}}$. Subsequently, two-layer FCs are utilized to unify the dimensions of the vector to the $\mathbb{R}^{d_{\text {model }}}$. The historical $P^o$ timesteps are encoded and represented as $\mathbb{R}^{P^o \times d_{\text {model }}}$.

To acquire time-varying road indicators, the summation method in \cite{zheng2020gman} is leveraged to fuse the above spatial embedding and temporal embedding as spatio-temporal embeddings (STEs). We denote the STEs of all $N$ roads in the road network for $P^o$ timesteps as $\mathcal{STE}[:P^o]=(STE_1, STE_2, \ldots,STE_{P^o}) \in \mathbb{R}^{P^o \times N \times d_{\text {model }}}$, which involve road network graph structures and time information.

\textit{2) Time Filtering Layer}: Previous works \cite{zhang2019flow, guo2019attention} have proved that traffic speeds are seriously affected by several timesteps in recent hours, days and weeks while exhibiting little dependencies with others. With the aim of filtering out relevant timesteps to effectively capture both the long- and short-term temporal dependencies, we propose a time filtering array behind the feature embedding layer to select important hourly and daily/weekly periodic timesteps from the referenced time window. As the name suggests, the time filtering array adopts an array to record all the significant historical timesteps (e.g., recent, daily, and weekly period timesteps) based on the experience of previous research. For example, assuming a gap of one hour between two neighboring timesteps, the traffic speed feature $H^{t_f}$ of current timestep $t_f$ is related to $H^{t_f-1}$, $H^{t_f-2}$, $H^{t_f-3}$, $H^{t_f-4}$, $H^{t_f-24}$, $H^{t_f-48}$, $H^{t_f-168}$, $H^{t_f-336}$, which are denoted as [0, -1, -2, -3, -4, -24, -48, -168, -336], i.e., the recent 4 hours, at the same time of the recent 2 days in 2 weeks. STE is simultaneously screened by the time filtering array. Therefore, after obtaining traffic speed embedding $\mathcal{H}$ and spatio-temporal embedding $\mathcal{STE}[:P^o]$, we leverage time filtering array to select $P$ relevant timesteps from $P^o$ timesteps as the output of the time filtering layer. The selected traffic speed embedding is denoted as $\mathcal{H}_I^{E n c}=\left(H^{t_1}, H^{t_2}, \ldots, H^{t_{P}}\right) \in \mathbb{R}^{P \times N \times d_{\text {model }}}$. STEs can separately be expressed as $\mathcal{STE} [:P]=(STE_1, STE_2, \ldots,STE_{P}) \in \mathbb{R}^{P \times N \times d_{\text {model }}}$, in which $ste_{i, t_j} \in \mathbb{R}^{d_{\text {model }}}$ represents the embedding of road $i$ at timestep $t_j$. The benefits of the time filtering array can be summarized as follows: 1) making the network eliminate the unimportant timesteps, i.e., those historical timesteps that have little or no correlation to traffic speeds at the predicted timestep, to reduce noise and 2) decreasing the computational cost by reducing the feature dimension.

To elucidate this idea, the processing procedure of the time filtering array is shown on the left of Fig. \ref{fig_4}. The traffic speed embedding at timestep $t_j$ is assumed to $H^{t_j}$, in which $h_i^{t_j}$ denotes the embedding of road $i$. For the target road $1$ and current timestep $t_0$, we determine that historical timesteps $t_2$, $t_i$, and $t_p$ affect $t_0$ based on the time filtering array. In addition, road $1$ is also affected by its current traffic speed. Thus, we filter out relevant timesteps, which are denoted as $\left(h_1^{t_0}, h_2^{t_0}, h_3^{t_0}, h_1^{t_2}, h_2^{t_2}, h_3^{t_2}, h_1^{t_i}, h_2^{t_i}, h_3^{t_i}, h_1^{t_p}, h_2^{t_p}, h_3^{t_p}\right)$.


\textit{3) Spatial-Temporal Encoding (ST-Encoding)}:
As previously mentioned, traffic speed fluctuations are affected by several factors. Specifically, in the spatial dimension, spatial dependencies have an impact on traffic speed fluctuations. For instance, a large influx of vehicles from neighboring roads leads to congestion and a dramatic decrease in the traffic speed on the target road. In the temporal dimension, traffic speed fluctuations are affected by long-term and short-term temporal dependencies. For example, traffic speed is influenced in a short term by factors such as traffic signals and accidents. Meanwhile, traffic speeds may indicate periodic peaks and valleys during specific times on a daily or weekly basis. Thus, to capture traffic speed fluctuations, it is essential to consider spatial dependencies and long-term and short-term temporal dependencies in traffic speed data, accordingly further enhancing prediction accuracy.

Up till now, we can acquire the traffic speed embedding and STEs of the relevant $P$ timesteps filtered by the previous time filtering layer. Then, ST-Encoding is presented to capture long-term and short-term temporal dependencies and spatial dependencies from these selected features. As shown in Fig. \ref{fig_4}, ST-Encoding consists of three components, namely spatial attention, temporal attention, and ST-Fusion. Spatial and temporal attention take a parallel structural design. Spatial attention is designed to extract spatial dependencies. Temporal attention is adopted to capture long-term and short-term temporal dependencies. In the end, according to the importance of the spatial information extracted by spatial attention and the temporal information captured by temporal attention, we develop an ST-Fusion to adaptively fuse each part.

\textit{\textbf{Spatial Attention}}: To extract the spatial dependencies, spatial attention based on a multi-layer graph attention network is proposed. Leveraging an attention mechanism, such a method can dynamically allocate weights to roads in the road network based on correlations between the roads.
 
In detail, given the filtered traffic speed embeddings $\mathcal{H}_I^{E n c} \in \mathbb{R}^{P \times N \times d_{\text {model }}}$ and STEs $\mathcal{STE} [:P] \in \mathbb{R}^{P \times N \times d_{\text {model }}}$ as the inputs of the multi-layer graph attention network, the final spatial representations output by the network are $\mathcal{H S}^{E n c} \in \mathbb{R}^{P \times N \times d_{\text {model }}}$. In the $l$th layer, its output is $\mathcal{H S}_{l}^{E n c} \in \mathbb{R}^{P \times N \times d_{\text {model }}}$, in which the representations of road $i$ at timestep $t_j$ represent $h s_{i, t_j, l}^{E n c} \in \mathbb{R}^{d_{\text {model }}}$.

For road $i$ at timestep $t_j$, the attention coefficient $\alpha_{i, v}^m$ between roads $i$ and $v$ at the $l$th layer is
\begin{equation}
\alpha_{i, v}^m=\frac{\exp \left(\varepsilon_{i, v}^m\right)}{\sum_{r=1}^V \exp \left(\varepsilon_{i, r}^m\right)},
\end{equation} where $V$ is the set of all roads. $\varepsilon_{i, v}^m$ denotes the relevance between roads $i$ and $v$ in the $m$th head and can be calculated by taking the inner product of the key vector of $v$ and the query vector of $i$, i.e.,
\begin{equation}
\varepsilon_{i, v}^m=\frac{\left\langle f_q^m\left[h s_{i, t_j, l-1}^{E n c}, s t e_{i, t_j}\right], f_k^m\left[h s_{v, t_j, l-1}^{E n c}, s t e_{v, t_j}\right]\right\rangle}{\sqrt{d}},
\end{equation} where $f_q^m$ and $f_k^m$ are Eq. (9) at the $m$th head attention of the query and key vector. $[*, *]$ and $\langle *, *\rangle$ denote binary concatenation and inner product, respectively.


After acquiring $\alpha_{i, v}^m$, the spatial representations $h s_{i, t_j, l}^{E n c} \in \mathbb{R}^{d_{\text {model }}}$ are formulated as:
\begin{equation}
h s_{i, t_j, l}^{E n c, m}=\sum_{r}^V \alpha_{i, r}^m f_v^m\left[h s_{r, t_j, l-1}^{E n c}, s t e_{r, t_j}\right],
\end{equation}
\begin{equation}
h s_{i, t_j, l}^{E n c}=\mathrm{BN}\left(\|_{m=1}^{\mathrm{M}}h s_{i, t_j, l}^{E n c, m} \mathrm{~W}_s+h s_{i, t_j, l-1}^{E n c}\right),
\end{equation} where $f_v^m$ is Eq. (9) for calculating the $m$th head attention of the value vector. $||$ and $M$ represent the concatenation and total headcount, respectively. BN is the Batch Normalization. After multiple-layer iterations using Eqs. (10)-(13), the final output of spatial representations in the road $i$ at timestep $t_j$ is computed as $h s_{i, t_j}^{E n c} \in \mathbb{R}^{d_{\text {model }}}$ and $h s_{i, t_j}^{E n c} \in \mathcal{H S}^{E n c}$. Therefore, spatial attention outputs spatial representations of the entire road network at historical $P$ timesteps, indicated as $\mathcal{H S}^{E n c} \in \mathbb{R}^{P \times N \times d_{\text {model }}}$.

\textit{\textbf{Temporal Attention}}: To effectively capture long-term and short-term temporal dependencies, we propose temporal attention with a multi-layer multi-head self-attention inspired by the Transformer \cite{vaswani2017attention}. Similar to spatial attention, this structure can dynamically allocate weights to different timesteps according to their significance.

Given only filtered traffic speed embeddings $\mathcal{H}_I^{E n c} \in \mathbb{R}^{P \times N \times d_{\text {model }}}$ as input of multi-layer multi-head self-attention, the final long-term and short-term temporal representations output by the structure are denoted as $\mathcal{H T}^{E n c} \in \mathbb{R}^{P \times N \times d_{\text {model }}}$. At the $r$th layer, its output is expressed as $\mathcal{H T}_{r}^{E n c} \in \mathbb{R}^{P \times N \times d_{\text {model }}}$, in which the representations of the road $i$ at timestep $t_j$ stand for $h t_{i, t_j, r}^{E n c} \in \mathbb{R}^{d_{\text {model }}}$.

For the road $i$ at timestep $t_j$, the attention coefficient $\mu_{t_{j, t}}^m$ on $t_j$ to $t$ in the $m$th head attention is
\begin{equation}
\mu_{t_j, t}^m=\frac{\exp \left(\rho_{t_j, t}^m\right)}{\sum_{t_r}^{\mathcal{N}_{t_{\mathrm{P}}}} \exp \left(\rho_{t_j, t_r}^m\right)},
\end{equation} where a set of historical timestep is represented as $\mathcal{N}_{t_{\mathrm{P}}}$; $\rho_{t_j, t}^m$ denotes the correlations between $t_j$ and $t$, which can be acquired by the inner product of the query vector of the timestep $t_j$ and the key vector of the timestep $t$:

\begin{equation}
\rho_{t_j, t}^m=\frac{\left\langle f_q^m\left(h t_{i, t_j, r-1}^{E n c}\right), f_k^m\left(h t_{i, t, r-1}^{E n c}\right)\right\rangle}{\sqrt{d}}.
\end{equation}

Once $\mu_{t_{j, t}}^m$ is obtained, the long-term and short-term temporal representations $h t_{i, t_j, r}^{E n c} \in \mathbb{R}^{d_{\text {model }}}$ are expressed as:
\begin{equation}
h t_{i, t_j, r}^{E n c, m}=\sum_{t_r}^{\mathcal{N}_{t_P}} \mu_{t_j, t_r}^m f_v^m\left(h t_{i, t_r, r-1}^{E n c}\right),
\end{equation}
\begin{equation}
h t_{i, t_j, r}^{E n c}=B N\left(\|_{m=1}^M h t_{i, t_j, r}^{E n c, m} W_t+h t_{i, t_j, r-1}^{E n c}\right).
\end{equation}

We use Eqs. (14)-(17) for multi-layer iterations. The final long-term and short-term temporal representations of the road $i$ at timestep $t_j$ are derived as $h t_{i, t_j}^{E n c} \in \mathbb{R}^{d_{\text {model }}}$ and $h t_{i, t_j}^{E n c} \in \mathcal{H T}^{E n c}$, respectively. Thus, temporal attention outputs long-term and short-term temporal representations of the entire road network at historical $P$ timesteps, denoted as $\mathcal{H T}^{E n c} \in \mathbb{R}^{P \times N \times d_{\text {model }}}$.

\textit{\textbf{ST Fusion}}: In order to adaptively integrate the acquired spatial and long-term and short-term temporal representations, we design a gating mechanism called ST-Fusion, as shown in Fig. \ref{fig_4}. Particularly, through a gating mechanism, the two types of representations are fused as:
\begin{equation}
z=\sigma\left(\left(\mathcal{H S}^{E n c} \odot \mathcal{H T}^{E n c}\right) W_{S T}^z+\mathcal{H T}^{E n c} W_{T}^z+b^z\right),
\end{equation}
\begin{equation}
\mathcal{H S T}^{E n c}=\mathcal{H S}^{E n c} \odot z+\mathcal{H T}^{E n c} \odot(1-z),
\end{equation} where $\mathcal{H S T}^{E n c} \in \mathbb{R}^{P \times N \times d_{\text {model }}}$ is the spatio-temporal representations as the output of ST Fusion, $z$ denotes the gating mechanism, $\sigma$ denotes the sigmoid activation, and $\odot$ represents the element-wise product. $W_{S T}^z \in \mathbb{R}^{d_{\text {model }} \times d_{\text {model }}}$, $W_T^z \in \mathbb{R}^{d_{\text {model }} \times d_{\text {model }}}$, and $b^z \in \mathbb{R}^{d_{\text {model }}}$ are the learnable parameters in ST Fusion.

\textit{4) Similar Attention Layer}: With ST-Encoding, we can acquire the representations of spatio-temporal correlations in the historical timesteps. The spatio-temporal correlations of future timesteps are different but similar to that of historical timesteps. Thus, to assist subsequent models in accurately capturing the spatio-temporal correlation of future timesteps and mitigate the adverse effects of error propagation in traffic speed prediction, we design similar attention to select the most relevant historical information from ST-Encoding. 
Intuitively, a region shares similar traffic speeds under similar STEs. 
Inspired by this, we first utilize the similarity in STEs to determine the degree of associations among historical and future timesteps. 
Since spatial embeddings are static and the timesteps to forecast belong to prior knowledge, we already know the future STEs. 
Based on the STEs of the history and future, a single-layer multi-head self-attention is employed to extract associations among historical and future timesteps. 
Second, relying on these associations, we select the most relevant historical spatio-temporal correlations for subsequent modeling. 
The specific implementation is as follows:

Let $\mathcal{H S T}^{E n c}=\left(H S T_{t_1}^{E n c}, H S T_{t_2}^{E n c}, \ldots, H S T_{t_P}^{E n c}\right) \in \mathbb{R}^{P \times N \times d_{\text {model }}}$, $\mathcal{H}_I^{D e c}=\left(H_{t_{P+1}}^{D e c}, H_{t_{P+2}}^{D e c}, \ldots, H_{t_{P+Q}}^{D e c}\right) \in \mathbb{R}^{Q \times N \times d_{\text {model }}}$ refer to the input and output of similar attention. The STEs of the historical $P$ and future $Q$ timesteps are denoted as $\mathcal{S T E}[:P]=\left(STE_{1}, STE_{2}, \ldots, STE_{P}\right) \in \mathbb{R}^{P \times N \times d_{\text {model}}}$ and $\mathcal{S T E}[P+1:P+Q]=\left(STE_{P+1}, STE_{P+2}, \ldots, STE_{P+Q}\right) \in \mathbb{R}^{Q \times N \times d_{\text {model }}}$, respectively.

First, the relevance $\tau_{t_a, t_b}$ between the future timestep $H_{t_a}^{D e c} \in \mathcal{H}_I^{D e c}$ and historical timestep $H S T_{t_b}^{E n c} \in \mathcal{H S T}^{E n c}$ is calculated, where $P<t_a \leq Q$, $t_b \leq P$. In the $m$th head attention, $\tau_{t_a, t_b}$ is present as:

\begin{equation}
\tau_{t_a, t_b}^m=\frac{\exp \left(\omega_{t_a, t_b}^m\right)}{\sum_{t_h=1}^{\mathcal{N}_{t_p}} \exp \left(\omega_{t_a, t_h}^m\right)},
\end{equation} where the attention coefficient $\omega_{t_a, t_b}^m$ between $t_a$ and $t_b$ is defined as follows:

\begin{equation}
\omega_{t_a, t_b}^m=\frac{\left\langle f_q^m\left({ } S T E_{t_a}\right), f_k^m\left({ } S T E_{t_b}\right)\right\rangle}{\sqrt{d}}.
\end{equation} Next, the output of similar attention $H_{t_a}^{Dec,m}$ on the $m$th head is acquired through aggregating the value vector of the historical timesteps with corresponding attention weights:

\begin{equation}
H_{t_a}^{D e c,m}=\sum_{t_r=1}^{\mathcal{N}_{t_P}} \tau_{t_a, t_r}^m f_v^m\left({ } H S T_{t_r}^{E n c}\right).
\end{equation} Then, the output of all heads is concatenated to yield the final result $H_{t_a}^{D e c}$,

\begin{equation}
{ } H_{t_a}^{D e c}=B N\left(\|_{m=1}^M H_{t_a}^{D e c,m} W_t\right),
\end{equation} where the unspecified definitions of parameters and operation symbols in Eqs. (20)-(23) are the same as before.

Finally, we obtain each $H_{t_a}^{D e c} \in \mathcal{H}_I^{D e c}$ for the ST-Decoding. The structure of ST-Decoding is the same as ST-Encoding. Hence, we omit the description of the ST-Decoding structure for simplicity. The long-term and short-term temporal and spatial representations in the future output by ST-Decoding are represented as $H S T^{D e c} \in \mathbb{R}^{Q \times N \times d_{\text {model }}}$.

\subsection{Prediction Layer Design}
This layer processes the final outputs of the STCL and SFPR modules to make traffic speed predictions for the road network. Particularly, to improve prediction accuracy, we integrate traffic diffusion and adaptation to traffic speed fluctuations as follows:

\begin{equation}
H^{\text {Pred }}=\alpha H S T^{\text {Dec }}+(1-\alpha) H^{S T C L},
\end{equation} where $\alpha$ is the learnable weight, $H^{S T C L} \in \mathbb{R}^{Q \times N \times d_{\text {model }}}$ is the output of STCL module, $H S T^{D e c} \in \mathbb{R}^{Q \times N \times d_{\text {model }}}$ is the output of SFPR module, and $H^{\text {Pred }} \in \mathbb{R}^{Q \times N \times d_{\text {model }}}$ denotes as the fused prediction features. An FC followed by Eq. (24) is applied to the prediction layer, gaining the predicted results according to $H^{\text {Pred }}$.

During the training process, the parameters of ICST-DNET can be optimized by minimizing the Mean Squared Error (MSE) loss function, presented as:
\begin{equation}
\text { loss }=\frac{\sum_{{t_s}=1}^Q \sum_{i=1}^N\left(x_{{t_s}, i}-\hat{y}_{{t_s}, i}\right)^2}{Q \times N}+\frac{\lambda}{2}\|W\|^2,
\end{equation} where $Q$ is the predicted series length, $N$ is the total number of target roads, $x_{{t_s}, i} \in X^{t_s}$ and $\hat{y}_{{t_s}, i} \in \hat{Y}^{t_s}$ denote observed and predicted values at the timestep $t_s$ on the road $i$. $\lambda$ is the regularization parameter, and $W$ is the learnable parameter of ICST-DNET.

\begin{table}[tpb]
\caption{Detailed statistics of METR-LA and Ningxia-YC\label{tab:table1}}
\centering
\begin{tabular}{llllll}
\hline
\textbf{Dataset} & \textbf{Nodes} & \textbf{Edges} & \textbf{Timesteps} & \textbf{Mean} & \textbf{Std} \\ \hline
METR-LA          & 207            & 1515           & 34272              & 53.71         & 20.26        \\
Ningxia-YC       & 108            & 458            & 8832               & 80.19         & 30.22        \\ \hline
\end{tabular}
\label{t2}
\end{table}
\subsection{Complexity Analysis}
We then analyze the complexity of ICST-DNET. According to the above definitions, let $N$, $P$, and $Q$ denote the number of roads in the road network, the filtered historical timesteps, and the predicted timesteps, respectively. In the STCL module, the complexity of the TCL layer is $\mathcal{O}\left(\mathrm{P} \times 2 \times d_{\text {model }}\right)$, where $d_{\text model }$ is a hyperparameter of ICST-DNET. SCL and NF layer incur complexities of $\mathcal{O}\left(2 \times d_{\text {model }} \times 2\right)$ and $\mathcal{O}\left(d_{\text {model }} \times 2 \times d_{\text {model }}\right)$, respectively. In CGG module, the time causality matrix and causal graphs remain $\mathcal{O}(1)$ for visualizing temporal and spatial causality. In the SFPR module, spatial and temporal attention take quadratic computational complexity w.r.t the number of roads $N$ and the length of timesteps $P$ and $Q$, respectively. Hence, for ST-Encoding, Its complexity is $\mathcal{O}\left(N^2 \times d_{\text {model }}+P^2 \times d_{\text {model }}+P \times N \times d_{\text{model }}^2\right)$, where $\mathcal{O}\left(P \times N \times d_{\text{model}}^2\right)$ represents the complexity of ST-Fusion. For similar attention, it has the complexity of $\mathcal{O}\left(Q \times P \times d_{\text {model }}\right)$. ST-Decoding necessitates a complexity of $\mathcal{O}\left(N^2 \times d_{\text {model }}+Q^2 \times d_{\text {model }}+Q \times N \times d_{\text{model}}^2\right)$, where $\mathcal{O}\left(Q \times N \times d_{\text {model}}^2\right)$ is the complexity of ST-Fusion. 

\section{EXPERIMENTS}\label{s4}

\subsection{Experimental Settings}
\textit{\textbf{Datasets}}: We evaluate ICST-DNET on two real-world traffic datasets METR-LA, released by \cite{li2017diffusion}, and Ningxia-YC, constructed by us \cite{zou2023will, zou2023novel}. In METR-LA, 207 sensors are utilized to collect traffic speed along the highways of Los Angeles County, spanning from March 1st, 2012, to June 30th, 2012. In Ningxia-YC, 108 sensors are employed to monitor traffic speed within Yinchuan City, China, covering the period from June 1st, 2021, to August 31st, 2021. The data collection frequencies for METR-LA and Ningxia-YC are 5 minutes and 15 minutes, respectively. Besides, Z-Score normalization is applied for data normalization. The two datasets are partitioned chronologically into training, validation, and test sets, with the radio of 7:1:2. Detailed statistical features of the two datasets are presented in TABLE \ref{t2}.


\begin{figure}[h]
    \centering
    \includegraphics[width=0.52\textwidth]{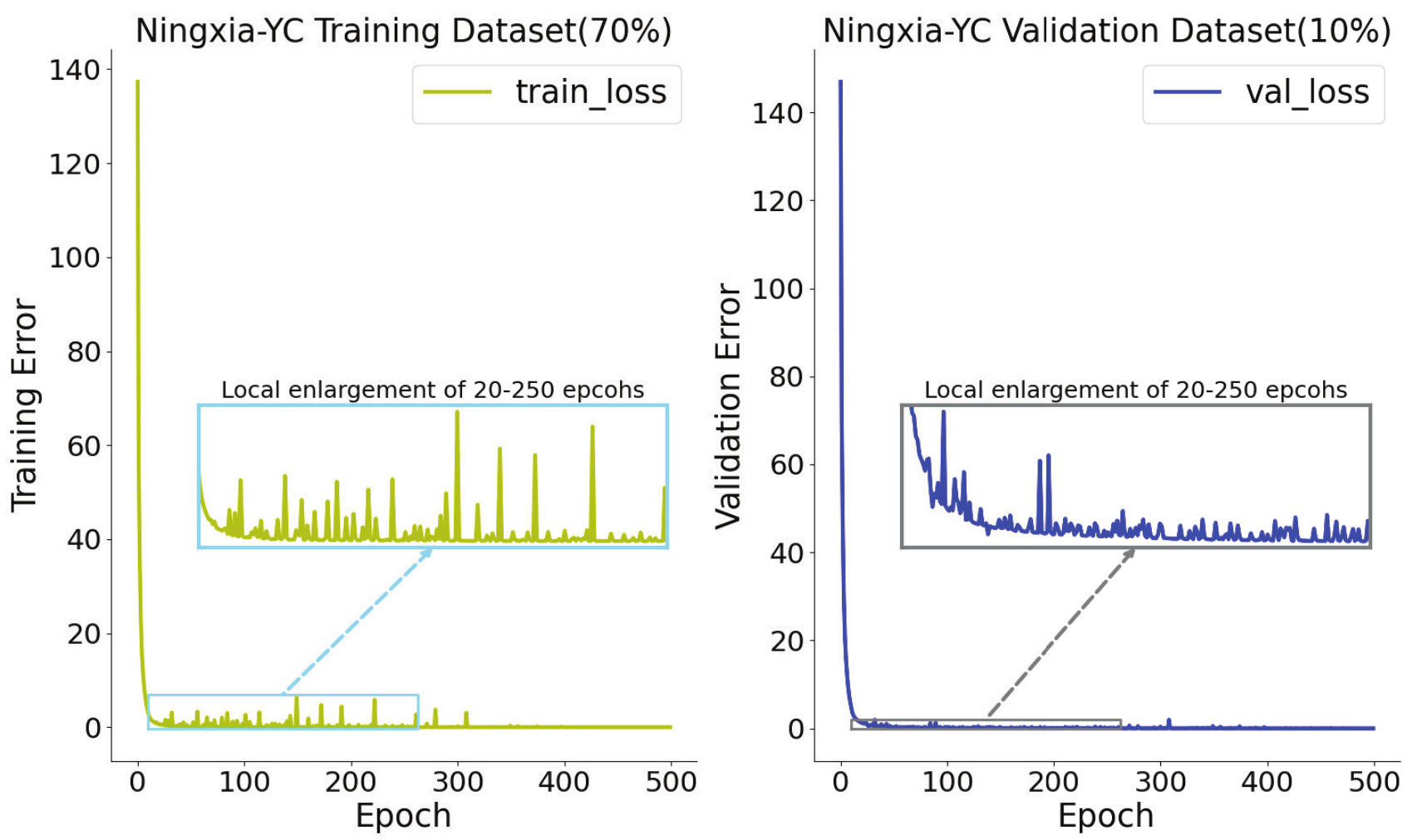}
    
    \caption{Training error (left) and validation error (right) in the Ningxia-YC. MAE is regarded as an error measure function.}
    \label{fig_5}
\end{figure}

\begin{table*}[t]
\caption{Traffic speed forecasting performance comparison of ICST-DNET and baselines on METR-LA and Ningxia-YC datasets}
\centering
\begin{tabular}{cccccccccccccc}
\hline
\multirow{2}{*}{dataset} & \multirow{2}{*}{Method} & \multicolumn{3}{c}{Horizon 3}           & \multicolumn{3}{c}{Horizon 6}           & \multicolumn{3}{c}{Horizon 12}                   & \multicolumn{3}{c}{Average}                      \\
                                  &                                  & MAE           & RMSE          & MAPE             & MAE           & RMSE          & MAPE             & MAE           & RMSE          & MAPE             & MAE           & RMSE          & MAPE             \\ \hline
\multirow{10}{*}{METR-LA}         & HA                               & 7.13          & 13.40         & 14.50\%          & 7.77          & 12.26         & 14.91\%          & 9.60          & 13.17         & 17.57\%          & 6.13          & 9.54          & 13.18\%          \\
                                  & SVR                              & 3.53          & 7.92          & 8.89\%           & 4.47          & 10.05         & 11.66\%          & 5.66          & 12.15         & 15.90\%          & 4.43          & 9.74          & 11.72\%          \\
                                  & LSTM                             & 3.44          & 6.30          & 9.60\%           & 3.77          & 7.23          & 10.90\%          & 4.37          & 8.69          & 13.20\%          & 3.86          & 7.41          & 11.23\%          \\
                                  & DCRNN                            & 2.77          & 5.38          & 7.30\%           & 3.15          & 6.45          & 8.80\%           & 3.60          & 7.59          & 10.50\%          & 3.17          & 6.47          & 8.87\%           \\
                                  & STGCN                            & 2.88          & 5.74          & 7.62\%           & 3.47          & 7.24          & 9.57\%           & 4.59          & 9.40          & 12.70\%          & 3.65          & 7.46          & 9.96\%           \\
                                  & GWN                              & \textbf{2.69} & \textbf{5.11} & \textbf{6.99\%}  & 3.11          & 6.20          & 8.48\%           & 3.58          & 7.37          & 10.33\%          & 3.13          & 6.23          & 8.60\%           \\
                                  & MTGNN                            & 2.69          & 5.18          & 6.86\%           & 3.05          & 6.17          & 8.19\%           & 3.49          & 7.23          & 9.87\%           & 3.08          & 6.19          & \textbf{8.31\%}  \\
                                  & ST-GRAT                          & 2.73          & 5.32          & 7.14\%           & 3.20          & 6.60          & 8.76\%           & 3.67          & 7.70          & 10.57\%          & 3.20          & 6.54          & 8.82\%           \\
                                  & GMAN                             & 2.77          & 5.48          & 7.25\%           & 3.07          & 6.34          & \textbf{8.35\%}  & \textbf{3.40} & 7.21          & \textbf{9.72\%}  & \textbf{3.08} & 6.34          & 8.44\%           \\
                                  & ICST-DNET                        & 2.75          & 5.32          & 7.08\%           & \textbf{3.07} & \textbf{6.14} & 8.42\%           & 3.45          & \textbf{7.09} & 9.90\%           & 3.09          & \textbf{6.18} & 8.46\%           \\ \hline
\multirow{10}{*}{Ningxia-YC}      & HA                               & 6.03          & 9.42          & 13.04\%          & 6.15          & 9.56          & 13.21\%          & 6.24          & 9.67          & 13.34\%          & 6.13          & 9.54          & 13.18\%          \\
                                  & SVR                              & 5.74          & 9.54          & 12.66\%          & 5.93          & 9.74          & 13.21\%          & 6.08          & 9.87          & 13.21\%          & 5.89          & 9.70          & 12.91\%          \\
                                  & LSTM                             & 5.77          & 9.43          & 20.07\%          & 6.15          & 9.62          & 20.38\%          & 6.57          & 10.23         & 20.98\%          & 6.16          & 9.76          & 20.48\%          \\
                                  & DCRNN                            & 5.28          & 8.86          & 11.92\%          & 5.35          & 8.95          & 11.98\%          & 5.46          & 9.07          & 12.22\%          & 5.35          & 8.94          & 12.01\%          \\
                                  & STGCN                            & 5.42          & 8.99          & 13.25\%          & 5.61          & 9.25          & 13.30\%          & 5.77          & 9.32          & 13.76\%          & 5.60          & 9.19          & 13.44\%          \\
                                  & GWN                              & 5.21          & 8.80          & 12.49\%          & 5.27          & 8.87          & 12.23\%          & 5.36          & 8.95          & 12.10\%          & 5.26          & 8.86          & 12.17\%          \\
                                  & MTGNN                            & 5.44          & 9.07          & 13.47\%          & 5.51          & 9.16          & 13.64\%          & 5.61          & 9.27          & 14.32\%          & 5.51          & 9.16          & 13.67\%          \\
                                  & ST-GRAT                          & 5.56          & 9.22          & 13.30\%          & 5.54          & 9.11          & 13.10\%          & 5.94          & 9.46          & 13.18\%          & 5.62          & 9.19          & 13.06\%          \\
                                  & GMAN                             & 5.32          & 8.93          & 12.76\%          & 5.33          & 8.95          & 12.68\%          & 5.40          & 9.03          & 12.76\%          & 5.35          & 8.97          & 12.74\%          \\
                                  & ICST-DNET                        & \textbf{5.12} & \textbf{8.79} & \textbf{11.32\%} & \textbf{5.22} & \textbf{8.83} & \textbf{11.45\%} & \textbf{5.31} & \textbf{8.91} & \textbf{11.90\%} & \textbf{5.21} & \textbf{8.84} & \textbf{11.56\%} \\ \hline
\end{tabular}
\label{t3}
\end{table*}

\textit{\textbf{Baselines}}: We compare our model with the following classic baselines:

\begin{itemize}
\item HA \cite{liu2004summary}: It simply adopts historical sequences to predict future traffic speeds.

\item SVR \cite{vanajakshi2004comparison}: The support vector based on the RBF kernel is utilized for iterative multi-step regression.

\item LSTM \cite{hochreiter1997long}: It is a variant of RNN that utilizes LSTM units to capture temporal dependencies and enable multi-step time series prediction.

\item DCRNN \cite{li2017diffusion}: It employs a bi-directional random walk on road networks to mine spatial dependencies and incorporates the Encoder-Decoder framework with GRUs to capture temporal dependencies.

\item STGCN \cite{yu2017spatio}: It adopts GCN to extract spatial dependencies and gated linear units to capture temporal dependencies.

\item GWN \cite{wu2019graph}: It is an enhanced version of STGCN. The model uses GCN with self-adaptive adjacency matrices to capture spatial dependencies. Furthermore, gated TCN rather than a gated linear unit is applied to mine temporal dependencies. 

\item MTGNN \cite{wu2020connecting}: It is the upgraded version of GWM and can obtain multi-scale spatio-temporal information.

\item ST-GRAT \cite{park2020st}: It employs the vanilla Transformer architecture and introduces novel spatial attention with diffusion graph matrix and additional sentinels to extract spatio-temporal correlations.

\item GMAN \cite{zheng2020gman}: It is an Encoder-Decoder framework whose encoder and decoder consist of spatial and temporal attention and gating mechanisms. Besides, GMAN develops a novel transform attention to avoid dynamic decoding in the vanilla Transformer.
\end{itemize}

\textit{\textbf{Metrics}}: The performance of different methods is evaluated through three metrics, namely Mean Absolute Error (MAE), Root Mean Squared Error (RMSE), and Mean Absolute Percentage Error (MAPE):
\begin{equation}
M A E=\frac{1}{Q \times N} \sum_{{t_s}=1}^Q \sum_{i=1}^N\left|x_{{t_s}, i}-\hat{y}_{{t_s}, i}\right|,
\end{equation}

\begin{equation}
R M S E=\sqrt{\frac{1}{Q \times N} \sum_{{t_s}=1}^Q \sum_{i=1}^N\left(x_{{t_s}, i}-\hat{y}_{{t_s}, i}\right)^2},
\end{equation}

\begin{equation}
M A P E=\frac{1}{Q \times N} \sum_{{t_s}=1}^Q \sum_{i=1}^N \frac{\left|x_{{t_s}, i}-\hat{y}_{{t_s}, i}\right|}{\hat{y}_{{t_s}, i}}.
\end{equation}

\textit{\textbf{Parameter Settings}}: In the experiment, the objective timestep $Q$ is set to 12. We predefine a time filtering array to select the most recent 12 timesteps and the recent 6 days and 2 weeks simultaneously, totaling 20 historical timesteps. The PyTorch library with two Tesla-A100 GPUs is employed for experiments. For training, the Adam optimizer is selected to minimize the MSE loss of 128 epochs on two datasets, and batch size and learning rate are set to 128 and 1e-3 separately. Additionally, $d_{\text {model }}$ is 32. In total, the hyperparameters in ICST-DNET include the number of attention heads $M$, the dimension of each head $d$, the NF layers $L_{N F}$, the spatial attention layers $L_S$, the temporal attention layers $L_T$, and the TCL residuals $L_{T C L}$. These parameters are fine-tuned on the validation set. After repeated experiments, it can be found that the best performing on the setting is $M=2, d=16, L_{N F}=3, L_S=1, L_T=2,$ and $L_{T C L}=3$.

For baselines, we implement LSTM with 64 hidden units and two layers. ST-GRAT is reproduced according to the parameters given in the paper. Other deep learning models are attained using the open source code. Additionally, statsmodels and sklearn packages in Python are employed to accomplish HA and SVR. The implementation codes of ICST-DNET and all baselines are open sourced on the GitHub homepage\footnote{https://github.com/ry19970812/ICST-DNET.git}.

\begin{table*}[t]
\caption{Ablation study of ICST-DNET on the Ningxia-YC dataset}
\centering
\begin{tabular}{cccccccccc}
\hline
$Q$                           & Metric    & Basic   & +TA     & +SA     & +STE    & +STF    & +SimA \& STD & +TFM    & ICST-DNET \\ \hline
\multirow{3}{*}{Horizon 3}  & MAE       & 28.06   & 5.92    & 5.67    & 5.46    & 5.43    & 5.29         & 5.31    & 5.12      \\
                            & RMSE      & 46.66   & 9.57    & 9.87    & 9.66    & 9.29    & 9.21         & 9.10    & 8.79      \\
                            & MAPE (\%) & 39.63\% & 14.58\% & 13.96\% & 14.97\% & 12.46\% & 13.39\%      & 10.46\% & 11.32\%   \\
\multirow{3}{*}{Horizon 6}  & MAE       & 28.07   & 5.99    & 5.83    & 5.52    & 5.45    & 5.41         & 5.35    & 5.22      \\
                            & RMSE      & 46.66   & 9.68    & 9.98    & 9.70    & 9.35    & 9.24         & 9.14    & 8.83      \\
                            & MAPE (\%) & 39.63\% & 13.94\% & 14.14\% & 15.90\% & 12.40\% & 12.40\%      & 10.45\% & 11.45\%   \\
\multirow{3}{*}{Horizon 12} & MAE       & 28.07   & 6.12    & 5.72    & 5.67    & 5.60    & 5.66         & 5.43    & 5.31      \\
                            & RMSE      & 46.65   & 9.86    & 10.16   & 9.63    & 9.45    & 9.36         & 9.26    & 8.91      \\
                            & MAPE (\%) & 39.63\% & 13.84\% & 14.39\% & 16.45\% & 12.51\% & 12.13\%      & 10.71\% & 11.90\%   \\
\multirow{3}{*}{Average}    & MAE       & 28.07   & 6.04    & 5.74    & 5.58    & 5.52    & 5.51         & 5.38    & 5.21      \\
                            & RMSE      & 46.66   & 9.74    & 10.04   & 9.65    & 9.39    & 9.29         & 9.19    & 8.84      \\
                            & MAPE (\%) & 39.63\% & 14.05\% & 14.22\% & 15.94\% & 12.47\% & 12.51\%      & 10.59\% & 11.56\%   \\ \hline
\end{tabular}
\label{t4}
\end{table*}

\begin{figure*}[t]
    \centering
    \includegraphics[width=0.8\textwidth, trim=0 110 0 0]{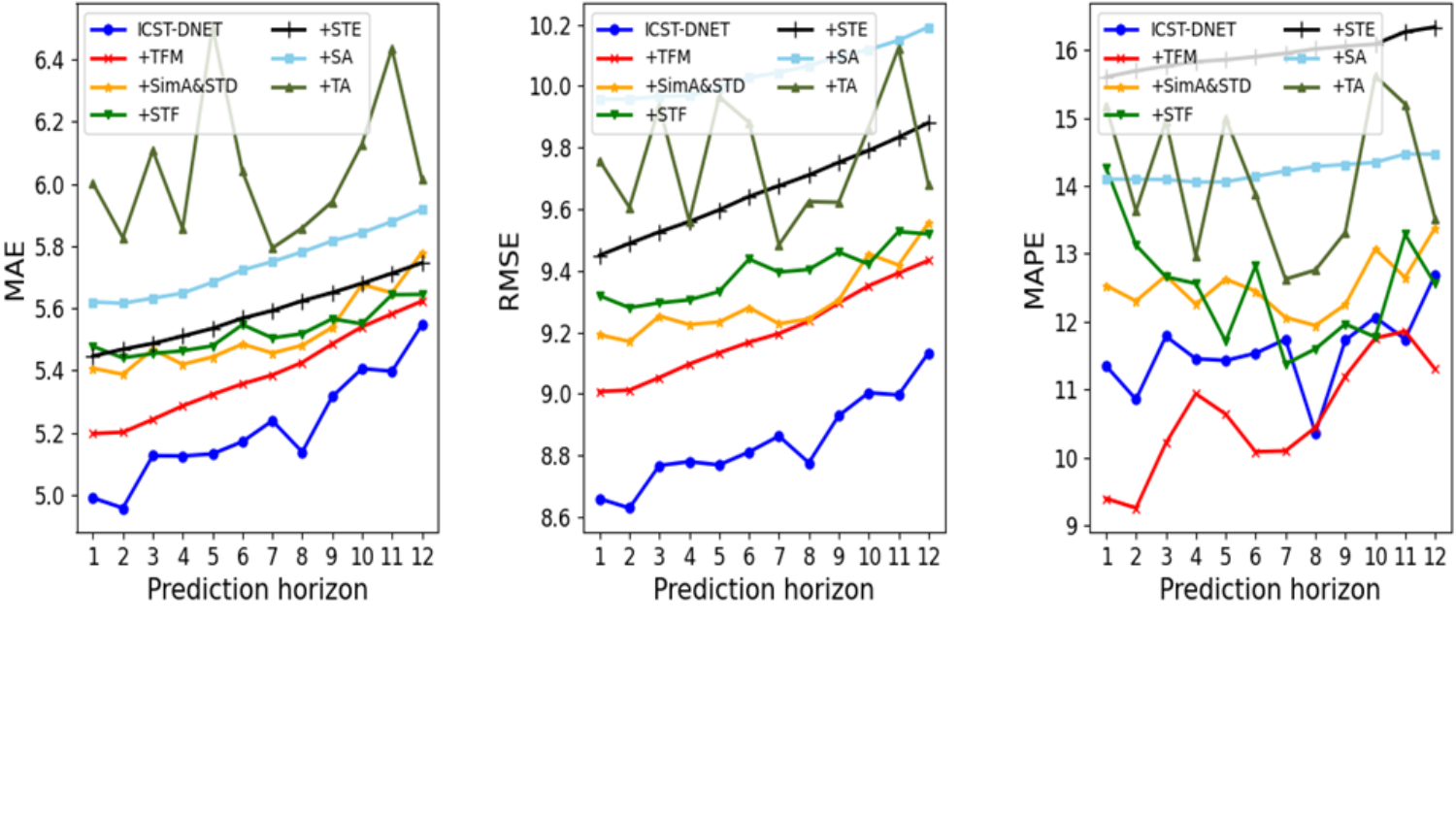}
    \caption{Curves composed of the results of each time slice for the ablation study on the Ningxia-YC dataset.}
    \label{fig_6}
\end{figure*}

\subsection{Experimental Results}

\textit{\textbf{The Convergence of ICST-DNET}}: To verify the stability of ICST-DNET, convergence curves are plotted on the dataset Ningxia-YC. As depicted in Fig. \ref{fig_5} (left), ICST-DNET converges rapidly in the training phase. In Fig. \ref{fig_5} (right), the validation loss fluctuates less when the training epochs reach 150. Based on these experimental results, two conclusions are summarized: (1) ICST-DNET achieves convergence within 150 epochs; (2) ICST-DNET exhibits excellent stability, ensuring minimal fluctuations on the validation set.

\textit{\textbf{Forecasting Performance Comparison}}:
The prediction precision of ICST-DNET and several state-of-the-art baselines for 3, 6, and 12 horizons ahead traffic speed prediction on METR-LA and Ningxia-YC datasets are contrasted in TABLE \ref{t3}. The following conclusions are drawn: 1) Due to the lack of high-order manual features, classical statistical and machine learning methods HA and SVR perform worse than the deep learning model LSTM. 2) The results of DCRNN and STGCN demonstrate that incorporating spatial networks into deep learning models accomplishes optimal performance. 3) The prediction accuracy of pre-defined GCN-based (i.e. build an adjacency matrix using the road network.) STGCN and DCRNN are inferior to adaptive GCN-based (i.e., back-propagate to train the adjacency matrix.) GWN and MTGNN, and self-attention-based GMAN and ST-GRAT. The commonality in the latter two groups of models lies in enhanced dynamic and global spatial receptive fields, which gather more information from distant yet similar roads.

ICST-DNET applies STCL and SFPR modules, which learn traffic diffusions within the road network and adapt to traffic speed fluctuations in multiple scenarios. This can improve prediction accuracy. Therefore, we observe that ICST-DNET achieves the best results on different horizons, especially on the Ningxia-YC dataset.

\textit{\textbf{Ablation Study}}: The following variants of ICST-DNET are designed to assess the effectiveness of each component proposed in our model:
\begin{itemize}
\item \textit{Basic} ICST-DNET removes STCL, CGG, and SFPR modules and only outputs results through two-layer FCs.

\item \textit{+TA} This model adds temporal attention to Basic.

\item \textit{+SA} The model joins spatial attention on +TA.

\item \textit{+STE} This model adds STEs on +SA.

\item \textit{+STF} This model adds ST-Fusion on +STE.

\item \textit{+SimA \& STD} This model equips +STF with similar attention and ST-Decoding.

\item \textit{+TFM} This model equips +SimA \& STD with time filtering layer.

\item \textit{ICST-DNET} Compared with +TFM, ICST-DNET adds STCL and CGG modules.
\end{itemize}

\begin{table*}[t]
\caption{The performance of ICST-DNET before and after replacing different modules on the Ningxia-YC dataset}
\centering
\begin{tabular}{cccccccccccccc}
\hline
\multirow{2}{*}{}                   & \multirow{2}{*}{Methods} & \multicolumn{3}{c}{Horizon 3} & \multicolumn{3}{c}{Horizon 6} & \multicolumn{3}{c}{Horizon 12} & \multicolumn{3}{c}{Average} \\
                                    &                          & MAE     & RMSE    & MAPE      & MAE     & RMSE    & MAPE      & MAE     & RMSE     & MAPE      & MAE    & RMSE    & MAPE     \\ \hline
\multirow{2}{*}{Temporal att.} & LSTM                     & 7.23    & 11.49   & 14.33\%   & 7.30    & 11.58   & 14.42\%   & 7.36    & 11.70    & 14.27\%   & 7.31   & 11.62   & 14.32\%  \\
                                    & GRU                      & 5.52    & 8.59    & 11.48\%   & 5.56    & 8.65    & 11.58\%   & 5.71    & 8.81     & 11.83\%   & 5.63   & 8.72    & 11.68\%  \\ \hline
\multirow{2}{*}{Spatial att.}  & CNN                      & 6.82    & 13.25   & 20.93\%   & 6.91    & 13.35   & 22.23\%   & 7.09    & 13.44    & 21.42\%   & 6.98   & 13.37   & 21.50\%  \\
                                    & GCN                      & 6.05    & 13.22   & 22.17\%   & 6.09    & 13.28   & 22.55\%   & 6.18    & 13.30    & 21.69\%   & 6.12   & 13.28   & 22.03\%  \\ \hline
\multirow{2}{*}{ST-Fusion}          & Gated Fusion \cite{zheng2020gman}    & 6.05    & 9.21    & 12.99\%   & 6.21    & 9.38    & 12.78\%   & 6.39    & 9.55     & 12.70\%   & 6.26   & 9.42    & 12.79\%  \\
                                    & F-Gate \cite{zou2023will}          & 5.49    & 8.68    & 14.18\%   & 5.63    & 8.83    & 14.08\%   & 5.81    & 9.02     & 13.53\%   & 5.69   & 8.89    & 13.83\%  \\ \hline
\multirow{3}{*}{SCL}                & VAR \cite{granger1969investigating}             & 7.02    & 10.30   & 13.38\%   & 7.15    & 10.56   & 13.69\%   & 7.32    & 10.82    & 14.07\%   & 7.16   & 10.56   & 13.71\%  \\
                                    & Copula \cite{bahadori2013examination}          & 6.01    & 9.03    & 12.29\%   & 6.23    & 9.14    & 12.38\%   & 6.54    & 9.37     & 12.67\%   & 6.26   & 9.18    & 12.45\%  \\
                                    & cLSTM \cite{tank2021neural}           & 5.65    & 8.97    & 12.38\%   & 5.96    & 9.09    & 12.26\%   & 6.19    & 9.22     & 12.53\%   & 5.93   & 9.09    & 12.39\%  \\ \hline
Ours                                & ICST-DNET                & 5.12    & 8.79    & 11.32\%   & 5.22    & 8.83    & 11.45\%   & 5.31    & 8.91     & 11.90\%   & 5.21   & 8.84    & 11.56\%  \\ \hline
\end{tabular}
\label{t5}
\end{table*}

For a fair comparison, ICST-DNET and all variants have the same settings, with only the above-mentioned differences. The estimated results in 3 horizons, 6 horizons, and 12 horizons are present in TABLE \ref{t4}, while the prediction accuracy evaluated for each timestep in the next 12 horizons on Ningxia-YC is displayed in Fig. \ref{fig_6}. Initially, since the basic model only focuses on linear dependencies, it results in a constant error at each timestep. In contrast to the basic model, dynamic temporal dependencies are introduced in +TA. Hence, the error is no longer fixed but follows an ascending curve. +SA simultaneously considers dynamic spatial dependencies, bringing about performance improvement over the base of +TA. In addition, the model +STE with STEs has achieved higher accuracy. Moreover, the results of +STF prove the effectiveness of the ST-Fusion. The model +SimA \& STD improves performance, thus determining the validity of the similar attention and ST-Decoding. The results of +TFM show that the time filtering array helps the model exclude some redundant historical timesteps, thus enhancing prediction accuracy. By comparing ICST-DNET and +TFM, we observe that learning traffic diffusions is essential for traffic speed prediction tasks.

\textit{\textbf{Module Replacement Analysis}}:
In this section, we investigate whether the various components within the ICST-DNET possess independence and substitutability, taking Ningxia-YC as an example. Several variants for components, such as temporal attention, spatial attention, ST-Fusion, and SCL layer, are constructed for experiments. The results are displayed in TABLE \ref{t5}.

Specifically, for temporal attention, we replace our Transformer-based architecture with other time series networks, such as LSTM and GRU. Regarding spatial attention, we attempt to replace the graph attention network with CNN and GCN. As for ST-Fusion, we replace it with the gating mechanism Gated Fusion in \cite{zheng2020gman} and F-Gate in \cite{zou2023will}. For SCL, the classical methods VAR \cite{granger1969investigating}, Copula \cite{bahadori2013examination}, and cLSTM \cite{tank2021neural} are adopted to replace it. TABLE \ref{t5} illustrates the results for these variants. Specific analyses of these variants are described below.

For temporal attention, GRU outperforms LSTM because of GRU's lightweight structure. Besides, the Transformer structure is superior to GRU under most evaluation metrics, indicating self-attention is more suitable for extracting long-term and short-term temporal dependencies than LSTM and GRU.

For spatial attention, we observe that CNN performs worse than GCN. It implies that CNN cannot effectively capture spatial dependencies between roads in the road network. In contrast, our method can dynamically compute the weights between roads, which significantly improves the prediction accuracy.

For ST-Fusion, F-Gate performs better than Gated Fusion. It shows that the element-wise multiplication used by F-Gate is more effective than the element-wise addition employed by Gated Fusion when fusing temporal and spatial features. Our ST-Fusion makes significant improvement compared with these two gates. One potential reason is that ST-Fusion combines the characteristics of both gates. Specifically, ST-Fusion first employs element-wise multiplication to fuse spatio-temporal features. It then incorporates the temporal representations through element-wise addition, emphasizing the importance of temporal dependencies for traffic speed prediction.

\begin{figure}[h]
    \centering
    \includegraphics[width=0.5\textwidth]{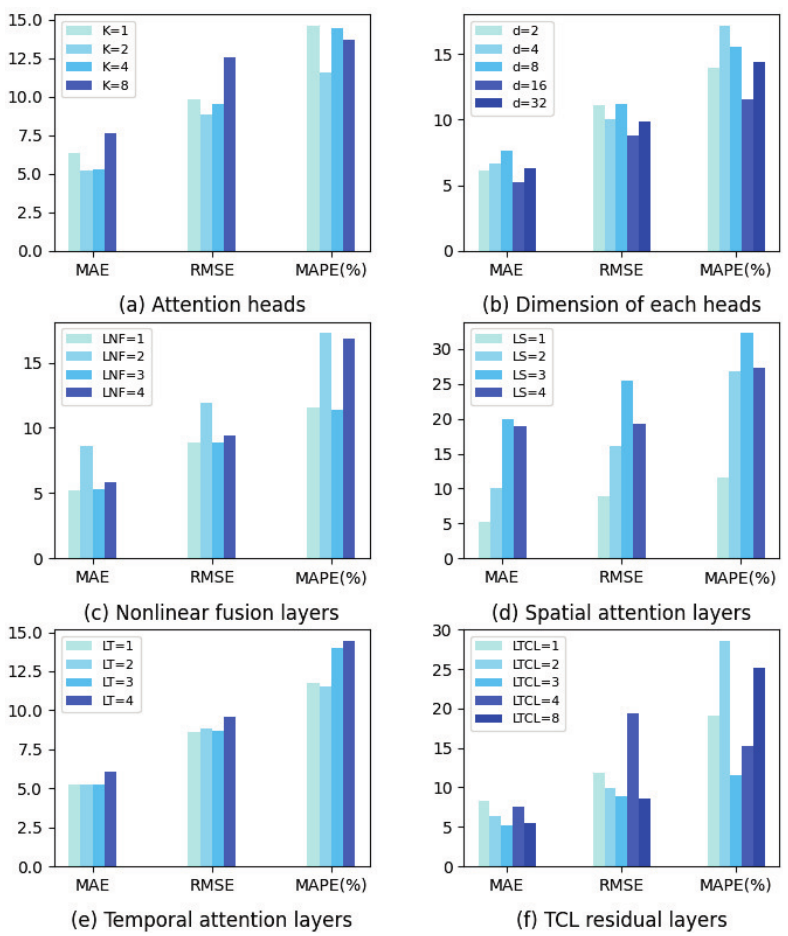}
    \caption{Experimental results under different hyperparameter settings.}
    \label{fig_7}
\end{figure}

\begin{figure*}[t]
    \centering
    \includegraphics[width=0.7\textwidth]{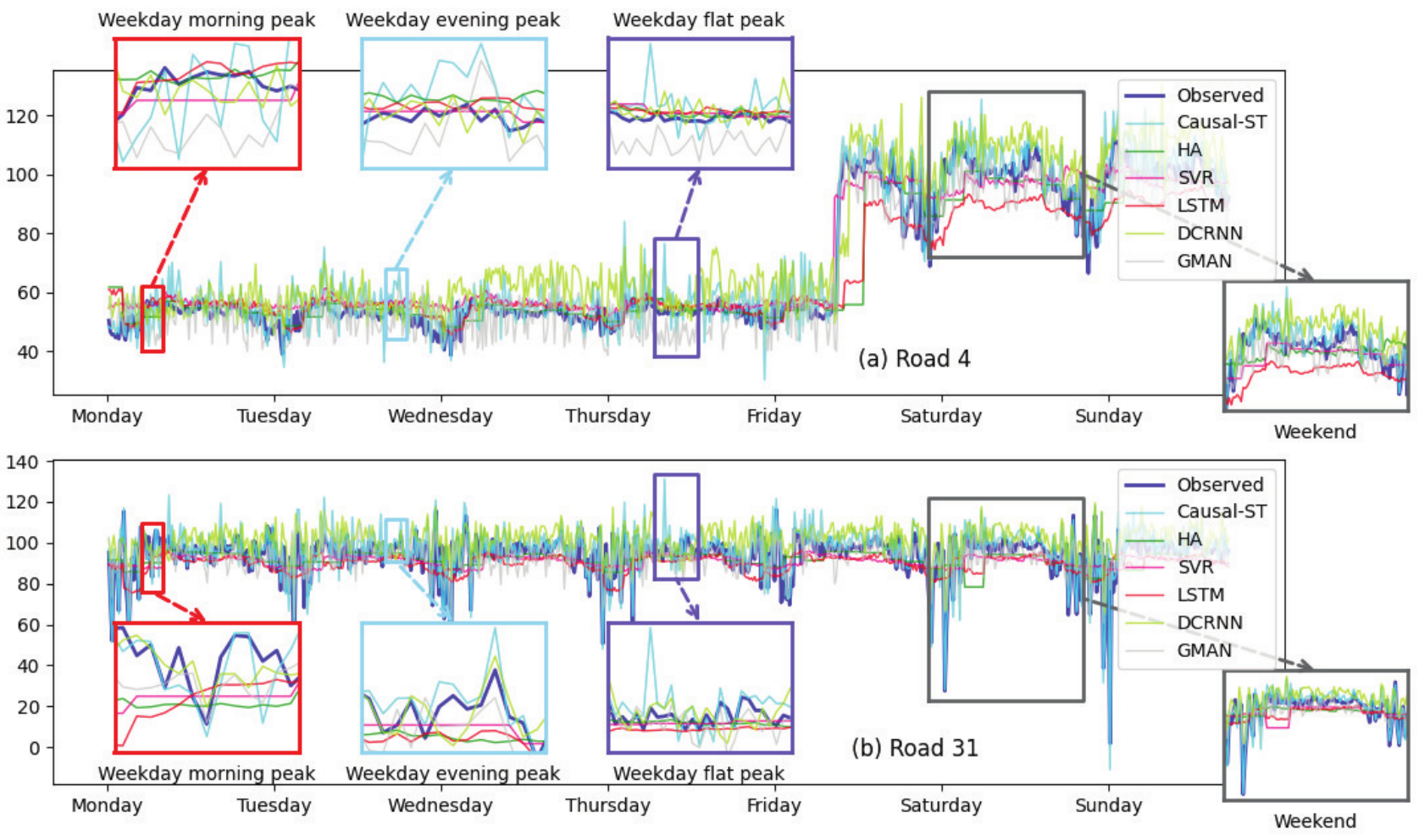}
    \caption{Case study of traffic speed prediction in multiple scenarios.}
    \label{fig_8}
\end{figure*}

For the SCL layer, we can see that VAR and Copula are worse than other methods mainly because of linearity. In contrast, cLSTM performs better. A significant reason is that cLSTM detects nonlinear spatial causality. The SCL layer achieves further improvement. It shows that our designed architecture can more systematically capture spatial causality.

\textit{\textbf{Effect of Hyperparameters}}: The average prediction of ICST-DNET under various hyperparameter settings over the next 12 horizons prediction on the Ningxia-YC dataset is displayed in Fig.\ref{fig_7}. When modifying a hyperparameter, the remaining hyperparameters retain their default optimal values ($M=2$, $d=16$, $L_{NF}=3$, $L_S=1$, $L_T=2$, $L_{TCL}=3$). The method to obtain optimal hyperparameters is bayesian optimization \cite{rasmussen2003gaussian}. In detail, We start by specifying a range of possible values for each hyperparameter. Then, a set of hyperparameters is randomly selected to evaluate prediction accuracy. Based on evaluation results, we dynamically adjust the search space to find the optimal hyperparameters. As shown in Figs. \ref{fig_7}(a), (b), (c), (e), and (f), we can see that the smaller the model, the easier it is to overfit, and the larger the model, the easier it is to underfit. Conversely. Fig. \ref{fig_7}(d) shows that a lower dimensional head achieves higher precision. This is because setting higher dimensional heads increases the accumulative errors.


\textit{\textbf{Case Study in Multiple Scenarios}}: In this section, we conduct a case study to intuitively show the adaptability of traffic speed for ICST-DNET in different scenarios, including morning rush hour (Monday 7:00-10:00), evening rush hour (Tuesday 16:00-19:00), flat peak hour (Thursday 10:00-16:00) in weekdays, and weekend (Saturday 9:00-21:00). We select road 4 and 31 in the Ningxia-YC dataset as the research objective and plot the fitted curves of traffic speed of last week based on HA, SVR, LSTM, DCRNN, GMAN and ICST-DNET in Fig. \ref{fig_8}.

For the morning and evening rush hours, changes in traffic speeds are non-stationary, i.e., traffic speed fluctuates more dramatically. Due to the limitations of modeling spatial and long-term and short-term dependencies, non-deep learning methods are less effective on non-stationary data. Compared with HA and SVR, the deep learning method LSTM has significant improvement in uncovering long-term and short-term temporal dependencies, i.e., the fit of the prediction changes from a straight line to a curve. In contrast to LSTM, DCRNN and GMAN improve in fitting the fluctuations. One main reason is that these networks capture spatial dependencies in addition to long-term and short-term temporal dependencies. We observe that ICST-DNET is more adaptable to traffic speed with more dramatic fluctuations than the other methods. It reveals that ICST-DNET can utilize the time filtering array to filter out pertinent historical timesteps. Spatial and long-term and short-term temporal dependencies are then extracted from selected timesteps.

\begin{figure}[h]
    \centering
    \includegraphics[width=0.5\textwidth]{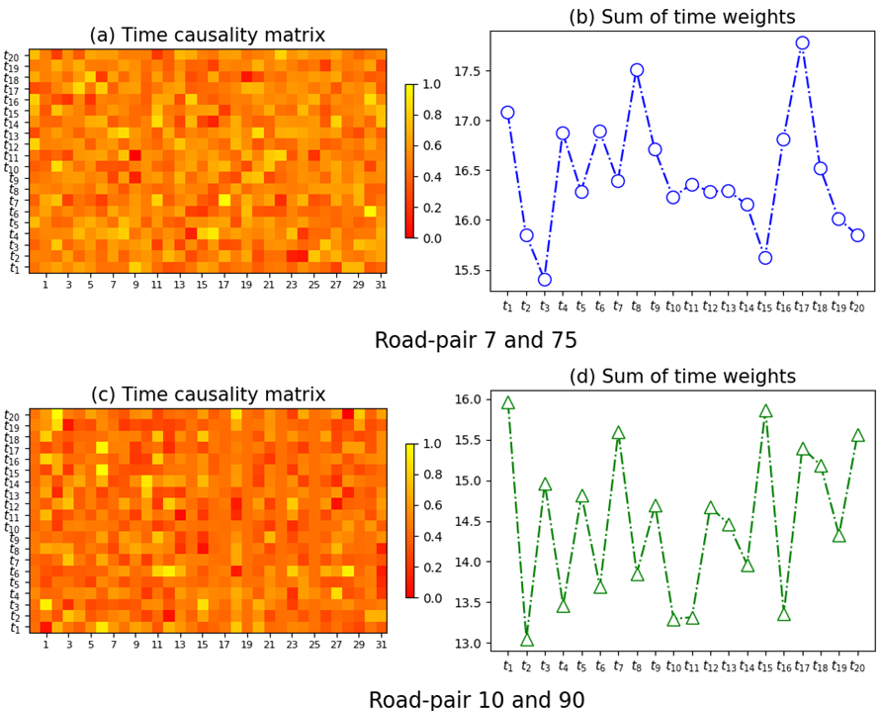}
    \caption{Temporal causality analysis on Ningxia-YC dataset. Figures (a) and (c) visualize the time causality matrix of the first residual block in the TCL layer. The y-axis represents the historical 20 horizons, and the x-axis represents the $d_{\text {model }}=32$. Figures (b) and (d) are the sum of weight dimensions for each historical horizon.}
    \label{fig_9}
\end{figure}

For flat peak hours on weekdays, traffic speeds change more gently, called stationarity. Owing to the restricted spatio-temporal feature mining capability, the fittings of the non-deep learning methods HA and SVR remain straight lines. LSTM has some strength in capturing long-term and short-term temporal dependencies, manifested in the slight volatility of the fitted curve. Compared to LSTM, DCRNN and GMAN achieve more accurate fitting. This is because they can learn dynamic spatial dependencies simultaneously. In comparison, we see that ICST-DNET also recognizes subtle changes in the stationary series. However, the stability needs to be further improved.

For weekend hours, the regularity of traffic speed changes is more complex, containing stationarity and non-stationarity. Compared with baselines, two conclusions can be drawn: 1) ICST-DNET has better oscillatory properties to realize the fitting of high peaks and low valleys in the non-stationary sequence. 2) our proposed model simultaneously completes excellent fitting of the stationary phases. This proves that the design of the SFPR module is more adaptive to traffic speed fluctuations in different scenarios.


\begin{figure}[h]
    \centering
    \includegraphics[width=0.5\textwidth]{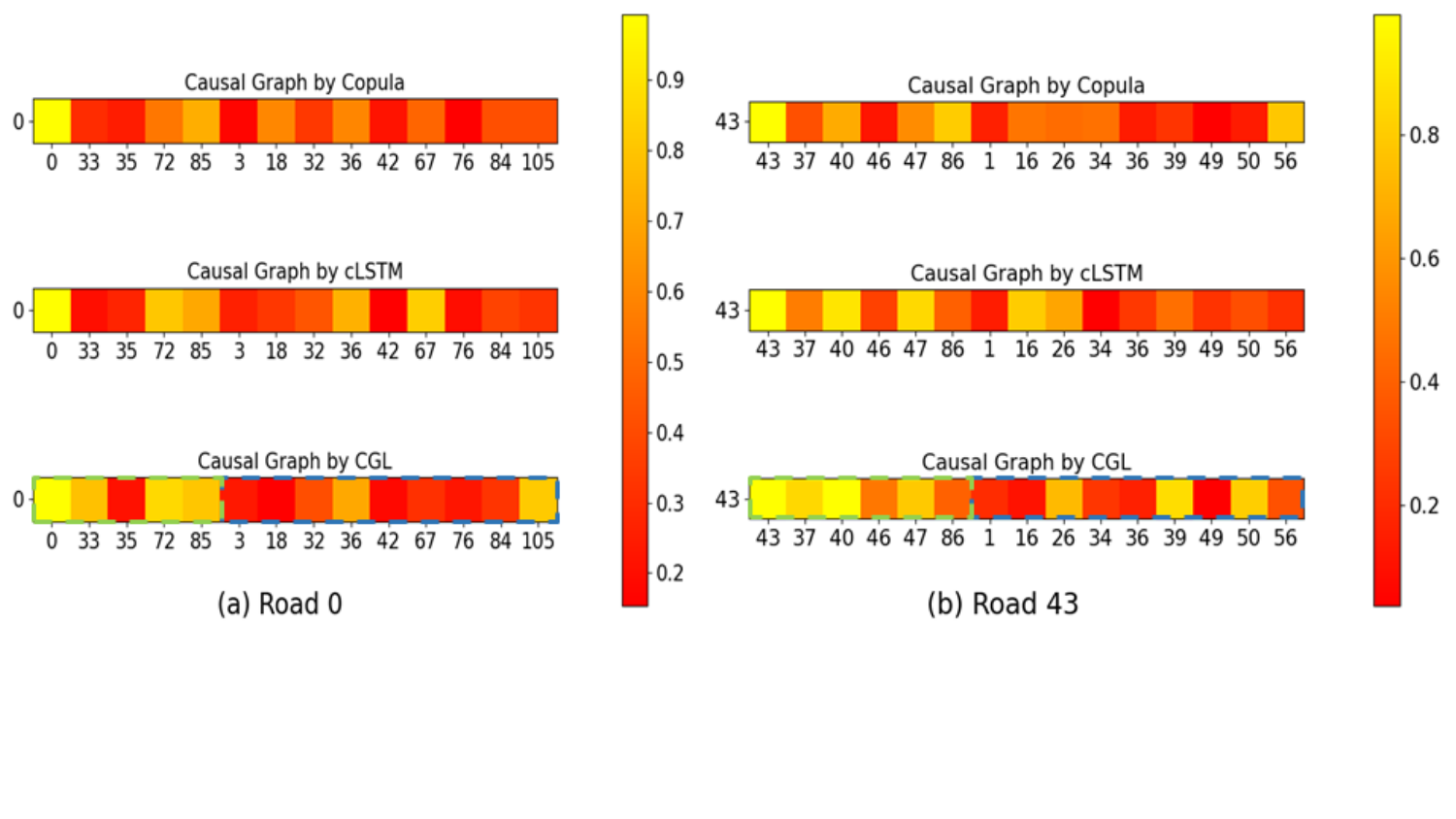}
    \vspace{-1.5cm}
    \caption{Learned causal graph of Ningxia-YC dataset. Green and blue boxes indicate first-order and second-order neighbors of the target roads, respectively.}
    \label{fig_10}
\end{figure}

\textit{\textbf{Interpretability Analysis}}: \textit{Temporal Causality Analysis}. Fig.9 visualizes the results of temporal causality extraction in the first residual of the TCL layer. We select road-pair 7 and 75 and road-pair 10 and 90 on the Ningxia-YC dataset for analysis. As shown in Figs. \ref{fig_9} (a) and (c), we observe that the TCL layer can learn temporal causality specific to each road. Furthermore, Figs. \ref{fig_9} (b) and (d) show that our proposed structure focuses on both recent (containing $t_1$, $t_8$ on road-pair 7 and 75, $t_1$, $t_7$ on road-pair 10 and 90) and long-term (containing $t_{1 7}$ on road-pair 7 and 75, $t_{1 5}$, $t_{2 0}$ on road-pair 10 and 90) temporal causality to each road. 

\textit{Spatial Causality Analysis}. This section details spatial causality by causal graphs on the Ningxia-YC dataset. Due to space limitation, we only visualize the spatial causality between the target roads 0 and 43 and their first- and second-order neighbors. The learned causal graphs by our proposed methods and baselines(e.g., Copula and cLSTM) are illustrated in Fig. \ref{fig_10}. We choose baselines based on the two best results in TABLE \ref{t5}. Causal graphs generated by ICST-DNET discover deterministic spatial causality relationships as follows: 1) the traffic speed of road 0 is subject to changes in its neighboring roads, which contain roads 33, 72, 85, 36, and 105; 2) the traffic speed on road 43 is also interrelated. The change of road 43 is triggered by the roads 37, 40, 47, 26, 39, and 50. Unlike our method, copula and cLSTM are unable to recognize more causal relationships between the roads. Furthermore, we can see that several second-order neighbors have causal associations with the target road. It indicates that indirect diffusions exist. Besides, the causality of the first-order neighbors is more prominent than that of the second-order ones. It proves that direct traffic diffusions have a more significant effect than indirect diffusions on traffic speed forecasting.

\section{CONCLUSION}\label{s5}
In this paper, we have proposed a novel framework, namely an interpretable causal spatio-temporal diffusion network (ICST-DNET) for traffic speed prediction. First, we have developed an STCL module to capture traffic diffusions through temporal and spatial causality discovery and correlate them with traffic speed forecasting. Second, based on the STCL module, we have designed a CGG module to interpret traffic diffusions. Expressly, from the temporal causality, a time causality matrix has been incorporated to visualize the temporal causality of past-to-future timesteps for the individual roads. From the spatial causality, causal graphs have been generated to simulate the physical processes of traffic diffusions in road pairs. Third, we have designed an SFPR module to adapt traffic speed fluctuations in multiple scenarios. Specifically, a predefined time filtering array has been introduced to filter out relevant historical timesteps. Spatial and long- and short-term temporal dependencies have been subsequently extracted from these selected timesteps. ICST-DNET has indicated effectiveness by evaluating on real datasets Ningxia-YC and METR-LA. For the future work, we will apply the proposed model to other spatio-temporal prediction tasks, such as air pollutant concentration forecasting.

\bibliographystyle{IEEEtran}
\bibliography{TITS}

\end{document}